\documentclass[10pt,twocolumn,letterpaper]{article}

\usepackage[pagenumbers]{cvpr} 

\definecolor{cvprblue}{rgb}{0.21,0.49,0.74}
\usepackage[pagebackref,breaklinks,colorlinks,allcolors=cvprblue]{hyperref}

\def\confName{CVPR}
\def\confYear{2027}

\def\b{\ensuremath\boldsymbol}

\usepackage{amsmath, amsthm} 
\usepackage{amsthm}

\newtheorem{proposition}{Proposition}
\newtheorem{lemma}{Lemma}

\usepackage{listings}
\usepackage{xcolor} 
\usepackage{multirow}
\title{Centroid Spectral Coupling: Implicit Alignment and Representation Diversity in Self-Supervised Learning}

\author{
M. Hadi Sepanj$^{1,*}$, Benyamin Ghojogh$^{2,*}$, Saed Moradi$^{1,*}$, Paul Fieguth$^{1}$ \\
\\
$^{1}$Department of Systems Design Engineering, University of Waterloo, Waterloo, ON, Canada \\
$^{2}$Department of Electrical and Computer Engineering, University of Waterloo, Waterloo, ON, Canada
}

\begin{document}
\maketitle

\begingroup
\renewcommand{\thefootnote}{\fnsymbol{footnote}}
\footnotetext[1]{Equal contribution.}
\endgroup

\begin{abstract}
Self-Supervised Learning (SSL) aims to learn representations
that align augmented views while preserving representation
diversity and preventing collapse. Existing non-contrastive
methods often address these objectives through multiple
explicitly weighted loss components. In this paper, we
introduce Centroid Spectral Coupling (CSC), a single-term
self-supervised learning objective that implicitly couples
augmented-view alignment, representation diversity, and
collapse resistance. CSC constructs centroids from pairs
of independently normalized augmented-view representations
and applies a ridge-based spectral objective to their
uncentered second-moment matrix. By preserving centroid
magnitudes, the proposed formulation encodes augmented-view
agreement while encouraging a balanced distribution of
representation energy across spectral directions.
We theoretically establish an exact decomposition of the
objective into nonnegative alignment and spectral-imbalance
components, characterize the geometry of global optima,
and derive finite-loss guarantees for augmented-view
alignment and representation diversity. We further
demonstrate that complete constant-output collapse is
a strict saddle in the normalized representation space.
Experiments on CIFAR-10, CIFAR-100, and Imagenette
demonstrate the effectiveness of CSC, achieving the
highest linear Top-1 accuracy among the evaluated
non-contrastive methods across all three benchmarks. A ridge-parameter sensitivity study further demonstrates
an empirical alignment--diversity trade-off: increasing
$\kappa$ strengthens augmented-view alignment while
reducing representation effective rank.
Moreover, CSC achieves the highest kNN Top-1 accuracy
among the evaluated non-contrastive methods
on both CIFAR benchmarks. These findings
demonstrate that a unified spectral objective can
produce competitive visual representations without
explicit contrastive negatives or separately weighted
alignment and diversity penalties.
\end{abstract}

\section{Introduction}
\label{sec:introduction}



Self-Supervised Learning (SSL) aims to learn transferable representations from unlabeled data by exploiting relationships between different views of the same sample~\cite{chen2020simclr,bardes2022vicreg}. A useful representation must satisfy three closely related requirements: (I) First, \textit{alignment} should make the representations of two augmented views of the same instance agree, encouraging invariance to nuisance transformations. (II) Second, \textit{representation diversity} should preserve differences across instances and spectral directions, providing the capacity to distinguish samples that may belong to different semantic classes. (III) Third, the objective must \textit{resist representation collapse}, in which all samples acquire an identical representation or variation is confined to too few dimensions. These requirements are interdependent: perfect positive-pair alignment alone admits a constant-output solution, whereas a diverse embedding need not preserve the correct correspondence between augmented views. The relationship between alignment and distributional spread has also been highlighted in analyses of contrastive learning~\cite{wang2020alignment}.

Existing SSL methods address these requirements in different ways. Contrastive methods such as SimCLR use positive and negative relationships in a contrastive objective~\cite{chen2020simclr}. Several prominent non-contrastive methods explicitly combine multiple objective components: VICReg adds separate invariance, variance, and covariance penalties~\cite{bardes2022vicreg}, while Barlow Twins combines diagonal cross-correlation matching with an off-diagonal redundancy-reduction penalty~\cite{zbontar2021barlow}. Other negative-free approaches, including BYOL and
SimSiam, avoid collapse through architectural asymmetry
and stop-gradient mechanisms
~\cite{grill2020byol,chen2021simsiam}.
Their training dynamics have also been studied
theoretically~\cite{tian2021ssl}.
Whitening-based SSL offers an alternative mechanism
for maintaining representation diversity
~\cite{ermolov2021whitening}. 
Related approaches further incorporate kernel-based dependence maximization \cite{li2021self,sepanj2025self}, spectral objectives derived from augmentation graphs \cite{haochen2021provable}, and information-theoretic  regularization of representation spectra \cite{kim2023vne}.
These approaches highlight different mechanisms for promoting agreement and representation diversity. This raises a natural question: \textit{can a single non-contrastive loss term simultaneously induce augmented-view alignment, encourage inter-instance diversity, and resist collapse, without explicitly adding separate alignment, variance, or covariance penalties?}

We propose \textit{Centroid Spectral Coupling} (CSC), a self-supervised objective that addresses these three goals through \textit{one ridge-spectral loss term}. For each instance, we independently normalize the projected representations of its two augmented views and form their arithmetic-mean centroid, \emph{without renormalizing that centroid}. We then apply a ridge effective-dimension functional to the uncentered second-moment matrix of the batch centroids. Crucially, the same spectrum contains information about both agreement and diversity. The magnitude of each centroid reflects agreement between its two views: misaligned views reduce centroid energy, while aligned unit-norm views maximize it. At the same time, the strict concavity of the ridge spectral response favors distributing this energy among multiple directions rather than concentrating it in a collapsed subspace. Hence, CSC couples alignment and diversity \emph{implicitly within a single scalar objective}; their appearance as separate nonnegative quantities in our theoretical decomposition is an analytical identity, not a prescription to optimize additional loss terms.


Prior work has explored several spectral objectives in self-supervised learning. These include connecting SSL to spectral embeddings and contrastive dualities~\cite{balestriero2022spectral,garrido2023duality}, as well as optimizing regularized trace criteria in predictive architectures~\cite{arbel2026peira}. Most closely related to our approach is Maximum Manifold Capacity Representations (MMCR)~\cite{yerxa2023mmcr}, which maximizes the nuclear norm of augmented-view centroids to implicitly promote alignment and uniformity~\cite{schaeffer2024towards}. However, a linear sum of singular values treats all spectral scales uniformly. It also lacks an explicit mechanism to control the trade-off between view agreement and spectral spread. We advance beyond nuclear-norm formulations by coupling classical ridge effective dimension~\cite{hsu2011ridge} directly with the uncentered second-moment matrix of unrenormalized augmented-view centroids. Because the ridge functional is strictly concave in the squared singular values, this coupling unlocks an exact analytical decomposition into nonnegative alignment and spectral-imbalance penalties. Moreover, it introduces a single hyperparameter ($\kappa$) that provably interpolates between spectral rank maximization and view alignment. This formulation enables a complete, assumption-free mathematical characterization. Specifically, we establish finite-batch certificates for alignment and centered variance, tight-frame global optimality, and strict-saddle instability of constant collapse, all without relying on unverifiable assumptions about latent semantic classes. Empirically, these theoretical advantages translate directly into practice. Controlled ablations confirm our predicted alignment-diversity dynamics, and CSC consistently outperforms MMCR and multi-term baselines in both linear probing and local kNN structure. 

Our principal contributions are threefold:
\begin{itemize}
    \item \textit{Implicit alignment from one spectral term.} We establish an exact relationship between centroid energy and augmented-view disagreement and derive a finite-batch alignment certificate. We also show by counterexample that renormalizing centroids can destroy this coupling, permitting zero modified loss despite severe disagreement between corresponding views.
    \item \textit{Implicit inter-instance diversity.} We derive an exact decomposition of CSC into nonnegative alignment-related and spectral-imbalance quantities, connect spectral balance to the centroid frame potential, and characterize the orthogonal or unit-norm tight-frame geometry of global minimizers. These results establish representation-level diversity without using class labels or an explicit diversity penalty.
    \item \textit{Mathematical resistance to collapse.} We prove a positive loss gap for rank-deficient batches, finite-loss certificates for centered rank and variance, and noncollapse at the global optimum. We further show that perfectly aligned constant-output collapse is a strict saddle in normalized representation space, while distinguishing this geometric property from an unconditional guarantee of escape under neural-network optimization.
\end{itemize}

\section{Centroid Spectral Coupling}
\label{sec:csc}

Self-supervised learning requires two complementary properties of a learned
representation. Different augmented views of the same sample should agree,
while representations of distinct samples should retain sufficient variation
to avoid collapse. The proposed \textit{Centroid Spectral Coupling} (CSC)
objective encourages both properties using one spectral functional. We first
establish an exact connection between augmented-view agreement and the total
energy of their centroids. We then show why energy maximization alone permits
collapse, motivating a ridge spectral functional which favors both energy and
its distribution across multiple directions. Finally, we distinguish the
resulting guarantees for augmented views and individual samples from stronger
claims about semantic classes, which are not observed during training.

\subsection{Augmented Views, Centroids, and Their Second Moment}
\label{subsec:csc_construction}

Consider a minibatch of $b$ unlabeled samples $\{\b{x}_i\}_{i=1}^b$. For each sample $\b{x}_i \in \mathbb{R}^d$, independently draw two augmentations $a_i$ and $a_i'$ and apply a shared encoder $f_\theta$ and projector $g_\phi$:
\begin{equation}
\b{z}_i=g_\phi(f_\theta(a_i(\b{x}_i))),\quad
\b{z}_i'=g_\phi(f_\theta(a_i'(\b{x}_i)))\in\mathbb{R}^p,
\label{eq:csc_raw}
\end{equation}
where $p$ is the dimensionality of the embedding space after the projector. 
We normalize the projected vectors individually:
\begin{equation}
\b{u}_i=\frac{\b{z}_i}{\max(\|\b{z}_i\|_2,\epsilon)},
\quad
\b{v}_i=\frac{\b{z}_i'}{\max(\|\b{z}_i'\|_2,\epsilon)},
\label{eq:csc_normalize}
\end{equation}
where $\epsilon=10^{-6}$.
Thus, $\|\b{u}_i\|_2,\|\b{v}_i\|_2\le1$, with equality when the corresponding raw vector has norm at least $\epsilon$. Unless otherwise stated, identities involving cosine similarity assume unit-norm views.

Define the augmented-view centroid:
\begin{equation}
\b{c}_i := \frac{\b{u}_i+\b{v}_i}{2}.
\label{eq:csc_centroid}
\end{equation}
This is the centroid of two views of the \textit{same instance}, not a class centroid. We use column-wise representations:
\begin{equation}\label{eq:csc_column_data}
\begin{aligned}
&\b{U}=[\b{u}_1,\ldots,\b{u}_b],\quad
\b{V}=[\b{v}_1,\ldots,\b{v}_b]\in\mathbb{R}^{p\times b},\\
&\b{C}=[\b{c}_1,\ldots,\b{c}_b]
=\frac{\b{U}+\b{V}}{2}\in\mathbb{R}^{p\times b}.
\end{aligned}
\end{equation}
Their uncentered second-moment matrix is:
\begin{equation}
\b{G} := \frac1b\b{C}\b{C}^{\top}
=\frac1b\sum_{i=1}^b\b{c}_i\b{c}_i^{\top}
\in\mathbb{R}^{p\times p}.
\label{eq:csc_second_moment}
\end{equation}
The matrix $\b{G}$ is symmetric and positive semidefinite. 

We define the total \textit{centroid energy} as:
\begin{equation}
t:=\operatorname{tr}(\b{G}).
\label{eq:csc_centroid_energy}
\end{equation}
Using the definition of $\b{G}$, the linearity of the trace, and the identity $\operatorname{tr}(\b{c}_i\b{c}_i^\top) =\b{c}_i^\top\b{c}_i$, we obtain:
\begin{align}
t
&=
\operatorname{tr}(\b{G})
=
\operatorname{tr}
\left(
\frac{1}{b}
\sum_{i=1}^{b}
\b{c}_i\b{c}_i^\top
\right)
\nonumber\\
&=
\frac{1}{b}
\sum_{i=1}^{b}
\operatorname{tr}
(\b{c}_i\b{c}_i^\top)
=
\frac{1}{b}
\sum_{i=1}^{b}\|\b{c}_i\|_2^2.
\label{eq:csc_trace_centroid_norm}
\end{align}
The centroid energy $t$ has a close relationship with augmented-view agreement; this relationship will be shown in Proposition~\ref{prop:csc_energy_alignment}.

\subsection{The Single Spectral Objective}
\label{subsec:csc_loss}

Let $r:=\min(b,p)$ denote the maximum possible rank
of the centroid matrix $\b{C}$ and its second-moment
matrix $\b{G}$. We denote the at most $r$ nonzero
eigenvalues of $\b{G}$, padded with zeros when needed,
by $\lambda_1,\ldots,\lambda_r\ge0$.
Since $\b{G}=(1/b) \b{C}\b{C}^\top$ is positive semidefinite
and $\operatorname{rank}(\b{G})\leq r=\min(b,p)$, its trace equals
the sum of its $r$ potentially nonzero eigenvalues:
\begin{equation}\label{equation_t_tr_G_sum_eig}
t=\operatorname{tr}(\b{G})=\sum_{j=1}^{r}\lambda_j.
\end{equation}
Moreover, since $\b{c}_i=(\b{u}_i+\b{v}_i)/2$ and
$\|\b{u}_i\|_2,\|\b{v}_i\|_2\leq1$, the triangle inequality gives:
\begin{equation*}
\|\b{c}_i\|_2
\leq\frac{\|\b{u}_i\|_2+\|\b{v}_i\|_2}{2}
\leq1.
\end{equation*}
Therefore:
\begin{equation}
0\leq \sum_{j=1}^r\lambda_j \overset{(\ref{equation_t_tr_G_sum_eig})}{=} t
\overset{(\ref{eq:csc_trace_centroid_norm})}{=}\frac{1}{b}\sum_{i=1}^{b}\|\b{c}_i\|_2^2
\leq1.
\end{equation}
Therefore, the sum of eigenvalues is bounded. 

We want to reward large total energy, but also avoid concentrating that
energy into only one or a few eigenvalues. For $\tau>0$, consider:
\begin{equation}
\begin{aligned}
&f_\tau(\lambda) := \frac{\lambda}{\lambda+\tau},
\label{eq:csc_f_tau}
\end{aligned}
\end{equation}
It is strictly monotonic because:
\begin{equation}
\begin{aligned}
f_\tau'(\lambda)=\frac{\tau}{(\lambda+\tau)^2}>0, \quad
f_\tau''(\lambda)=-\frac{2\tau}{(\lambda+\tau)^3}<0.
\label{eq:csc_concavity}
\end{aligned}
\end{equation}
Its strict monotonicity rewards spectral energy,  while strict concavity rewards a more even spectrum at fixed energy~\cite{marshall2011inequalities,bhatia1997matrix}. 


\subsubsection{CSC Loss Function}

We define the CSC loss function as:
\begin{equation}
\ell_{\mathrm{csc}}
=
1-\frac{1+\kappa}{r}
\sum_{j=1}^{r}
f_\tau(\lambda_j),
\label{eq:csc_loss_f}
\end{equation}
where $\kappa > 0$ is the \textit{ridge hyper-parameter}.
Set:
\begin{align}\label{equation_tau}
\tau:= \frac{\kappa}{r},
\end{align}
for $\kappa>0$. According to Eq. (\ref{eq:csc_f_tau}), the CSC loss function becomes:
\begin{equation}
\boxed{
\ell_{\mathrm{csc}}
=1-\frac{1+\kappa}{r}
\sum_{j=1}^r\frac{\lambda_j}{\lambda_j+\kappa/r}.}
\label{eq:csc_spectral_loss}
\end{equation}

\begin{proposition}\label{proposition_loss_csc_tr_G}
The CSC loss function is equivalent to:
\begin{equation}
\boxed{
\ell_{\mathrm{csc}}
=1-\frac{1+\kappa}{r}
\operatorname{tr}\!\left[
\b{G}(\b{G}+\tau\b{I}_p)^{-1}\right],}
\label{eq:csc_matrix_loss}
\end{equation}
where $\b{I}_p$ is the $p \times p$ identity matrix.
\end{proposition}

\begin{proof}
See supplementary material \ref{section_proof_loss_csc_tr_G}.
\end{proof}

The trace in Eq. (\ref{eq:csc_matrix_loss}) is the \textit{ridge effective dimension} (or \textit{effective degrees of freedom}) of $\b{G}$~\cite{wahba1990spline,caponnetto2007optimal,hsu2011ridge}, rather than the raw trace $t=\operatorname{tr}(\b{G})$. The prefactor normalizes the ideal score to one. The objective contains no additional explicit invariance, covariance, or contrastive term.
See supplementary material \ref{section_minimal_loss_function} for implementation of CSC loss function in PyTorch.

\begin{proposition}[Bounds and global optimum of CSC]
\label{prop:csc_optimum}
For $\kappa>0$, the CSC score, $s_\kappa := 1-\ell_{\mathrm{csc}}$, satisfies:
\begin{equation}
0\le s_\kappa= 1-\ell_{\mathrm{csc}}
\le\frac{(1+\kappa)t}{t+\kappa}\le1.
\label{eq:csc_score_upper}
\end{equation}
The first upper bound is attained if and only if
$\lambda_1=\cdots=\lambda_r=t/r$, while the second
is attained if and only if $t=1$.

Consequently, the CSC loss is bounded:
\begin{equation}
0\le\ell_{\mathrm{csc}}\le1.
\end{equation}
Both bounds are attainable. In particular, the
global minimum is $\ell_{\mathrm{csc}}=0$, attained
if and only if:
\begin{equation}
\begin{aligned}
&\b{u}_i=\b{v}_i,\quad
\|\b{u}_i\|_2=\|\b{v}_i\|_2=1,
\quad \forall i,\\
&\lambda_1=\cdots=\lambda_r=\frac{1}{r}.
\end{aligned}
\end{equation}
Thus, the global optimum requires both perfect
augmented-view alignment and a flat centroid spectrum.
\end{proposition}

\begin{proof}
See supplementary material \ref{section_proof_csc_optimum}.
\end{proof}

\subsubsection{Importance of Magnitude of Centroids}

It is important to avoid
renormalizing the centroids $\{\b{c}_i\}_{i=1}^b$ after averaging
the augmented views in Eq. (\ref{eq:csc_centroid}). The magnitude of each
centroid contains information about the
agreement between its corresponding views.
Renormalization removes this information and
can therefore weaken the alignment mechanism.
Indeed, as demonstrated by the counterexample
in supplementary material~\ref{app:csc_centroid_renormalization},
renormalized centroids can attain zero modified
CSC loss even when their corresponding augmented
views are substantially misaligned. Preserving
the original centroid magnitudes is therefore
essential to the coupling of alignment and
spectral diversity in CSC.

\subsubsection{Effect of the Ridge Parameter}

The ridge parameter $\kappa$ controls the relative
emphasis on spectral diversity and augmented-view
alignment. 

\begin{proposition}
For a fixed second-moment matrix
$\b{G}$, the CSC score
$s_\kappa=1-\ell_{\mathrm{csc}}$ satisfies:
\begin{equation}
\lim_{\kappa\to0^+}s_\kappa
=\frac{\operatorname{rank}(\b{G})}{r},
\qquad
\lim_{\kappa\to\infty}s_\kappa=t.
\label{eq:csc_kappa_limits}
\end{equation}
\end{proposition}

\begin{proof}
See supplementary material \ref{app:csc_kappa_limits}.
\end{proof}
Thus, smaller values of $\kappa$ increasingly
emphasize spectral rank, whereas larger values
emphasize centroid energy. As will be established
in Proposition~\ref{prop:csc_energy_alignment},
for unit-normalized augmented views, the centroid
energy satisfies $t=1- 1 / (4b) \|\b{U}-\b{V}\|_F^2$.
Consequently, maximizing centroid energy is
equivalent to minimizing the squared distance
between corresponding augmented views.
Therefore, larger values of $\kappa$ increasingly
emphasize augmented-view alignment, whereas
finite intermediate values allow CSC to couple
alignment with spectral diversity.

\subsection{Implicit Alignment}
\label{subsec:csc_within_instance}

The first theoretical property of CSC concerns the
agreement between augmented views of the same instance.
We show that the total centroid energy is directly related
to the squared distance between corresponding views.
This relationship also provides an exact interpretation
in terms of within-instance variance.

\begin{proposition}[Centroid energy and augmented-view agreement]
\label{prop:csc_energy_alignment}
Let $\b{u}_i,\b{v}_i\in\mathbb{R}^p$ denote the
two normalized representations of instance $i$, and let $c_i$, $\b{G}$, and $t$ be defined by Eqs. (\ref{eq:csc_centroid}), (\ref{eq:csc_second_moment}), and (\ref{eq:csc_centroid_energy}), respectively. 
If $\|\b{u}_i\|_2=\|\b{v}_i\|_2=1$ for all $i$,
then:
\begin{equation}
t=
\frac{1}{2}
+\frac{1}{2b}
\sum_{i=1}^{b}\b{u}_i^\top\b{v}_i
=
1-\frac{1}{4b}\|\b{U}-\b{V}\|_F^2,
\label{eq:csc_trace_distance}
\end{equation}
where $\b{U}$ and $\b{V}$ are defined in Eq. (\ref{eq:csc_column_data}).

Furthermore, define the average within-pair scatter as:
\begin{equation}
w_{\mathrm{pair}}
:=
\frac{1}{2b}
\sum_{i=1}^{b}
\left(
\|\b{u}_i-\b{c}_i\|_2^2+
\|\b{v}_i-\b{c}_i\|_2^2
\right).
\label{eq:csc_pair_scatter}
\end{equation}
Then:
\begin{equation}
w_{\mathrm{pair}}
=
\frac{1}{4b}\|\b{U}-\b{V}\|_F^2
=
1-t.
\label{eq:csc_pair_variance}
\end{equation}
Consequently, maximizing centroid energy $t$ is
equivalent to minimizing the within-instance scatter
of the normalized augmented views.
\end{proposition}

\begin{proof}
See supplementary material \ref{section_proof_csc_energy_alignment}.
\end{proof}

For the numerical normalization in Eq.~(\ref{eq:csc_normalize}), some representations may have norms smaller than one.
In this general case:
\begin{equation}
t=
\frac{\|\b{U}\|_F^2+\|\b{V}\|_F^2}{2b}
-
\frac{\|\b{U}-\b{V}\|_F^2}{4b}.
\label{eq:csc_trace_general}
\end{equation}

Since $\|\b{U}\|_F^2,\|\b{V}\|_F^2\le b$,
we obtain:
\begin{equation}
\frac{1}{b}\|\b{U}-\b{V}\|_F^2
\le4(1-t).
\label{eq:csc_trace_floor}
\end{equation}
Hence, even with normalization-floor rows, larger
centroid energy constrains the discrepancy between
the two augmented views.

Proposition \ref{prop:csc_energy_alignment} establishes an exact connection between centroid energy and within-instance variation. However, maximizing centroid energy $t$ alone is insufficient for preventing collapse.
Indeed, if all representations equal the same
unit vector $\b{w}$:
\begin{equation}
\b{u}_i=\b{v}_i=\b{w},
\quad
\|\b{w}\|_2=1,
\quad \forall i,
\end{equation}
then:
\begin{equation}
\b{G}=\b{w}\b{w}^\top,
\quad
t=1,
\quad
\operatorname{rank}(\b{G})=1.
\end{equation}
Therefore, perfect augmented-view agreement and
maximum centroid energy may coexist with complete
representation collapse. This illustrates why CSC
does not maximize the trace $t$ alone, but instead
couples centroid energy with spectral diversity.



\begin{proposition}[Decomposition of CSC loss into alignment and spectral-imbalance terms]
\label{prop:csc_decomposition}
Let $q:=t/r$, $\gamma:=(1+\kappa)/r$, and $\tau:=\kappa/r$. Then:
\begin{equation}
\ell_{\mathrm{csc}}
=\underbrace{\frac{\kappa(1-t)}{t+\kappa}}_{\ell_{\mathrm{align}}}
+\underbrace{\frac{\gamma\tau}{(q+\tau)^2}
\sum_{j=1}^r\frac{(\lambda_j-q)^2}{\lambda_j+\tau}}_{\ell_{\mathrm{spec}}}.
\label{eq:csc_exact_decomp}
\end{equation}
Both components are nonnegative. The second component vanishes if and only if all $r$ eigenvalues are equal.
\end{proposition}

\begin{proof}
See supplementary material \ref{section_proof_csv_decomposition}. 
\end{proof}

Thus, the CSC loss admits an exact decomposition into a
nonnegative centroid-energy penalty and a nonnegative
spectral imbalance penalty. 
Hence, the single CSC objective simultaneously encourages
agreement between corresponding augmented views and a
balanced distribution of centroid energy across spectral
directions.

According to Eq.~\eqref{eq:csc_pair_variance} for unit views, the $\ell_{\mathrm{align}}$ in Eq. (\ref{eq:csc_exact_decomp}) can be stated as:
\begin{equation}
\ell_{\mathrm{align}}
=\frac{\kappa w_{\mathrm{pair}}}
{1+\kappa-w_{\mathrm{pair}}}.
\label{eq:csc_pair_monotone}
\end{equation}
This expression increases strictly with $w_{\mathrm{pair}}$. Thus, the alignment component of the single CSC loss implicitly penalizes within-instance variation.

\begin{proposition}[Finite-batch augmented-view alignment certificate]
\label{prop:csc_align_bound}
If $\ell_{\mathrm{csc}}\le\eta<1$, then:
\begin{equation}
\frac{1}{b}\|\b{U}-\b{V}\|_F^2
\le\frac{4\eta(1+\kappa)}{\kappa+\eta}.
\label{eq:csc_align_cert}
\end{equation}
\end{proposition}

\begin{proof}
See supplementary material \ref{section_proof_csc_align_bound}.
\end{proof}

Proposition \ref{prop:csc_align_bound} shows that if the CSC loss is sufficiently small, then the two augmented views of each sample must, on average, be close to each other.
The alignment property established in
Proposition~\ref{prop:csc_align_bound} concerns two augmented
views of the same instance. Since common data augmentations
are typically designed to preserve semantic identity, reducing
the discrepancy between these views encourages representations
to be invariant to the applied transformations.

\subsection{Implicit Representation Diversity}
\label{subsec:csc_spectral_diversity}

Beyond augmented-view alignment, CSC encourages
diversity among instance representations implicitly by
favoring a balanced centroid spectrum.
As proved in supplementary
material \ref{app:csc_diversity_details}, for the centroid matrix $\b{C}$, we have:
\begin{equation}
\operatorname{tr}(\b{G}^2)
=
\frac{1}{b^2}
\sum_{i,j=1}^{b}
(\b{c}_i^\top\b{c}_j)^2
\geq \frac{t^2}{r},
\end{equation}
with equality precisely when the $r$ potentially
nonzero eigenvalues of $\b{G}$ are equal.
At fixed centroid energy $t$, the CSC objective
favors this flat spectrum, thereby reducing
aggregate squared similarities between instances. 

Furthermore, at zero CSC loss, the centroids
are orthogonal when $b\leq p$ and form a
unit-norm tight frame when $b>p$ (see proofs in supplementary
material \ref{app:csc_diversity_details}).
These properties establish spectral diversity
without requiring class labels or explicit
diversity regularization. The resulting geometry is closely related to classical finite normalized tight frames ~\cite{benedetto2003finite,waldron2018introduction} and achieves the classical Welch lower bound on aggregate squared inner products~\cite{welch1974lower}.

Detailed propositions, geometric characterizations,
and proofs for implicit representation diversity are provided in the supplementary
material \ref{app:csc_diversity_details}.

\subsection{Collapse Resistance of the CSC Loss}
\label{subsec:csc_collapse}

While providing implicit alignment and representation diversity, CSC implicitly discourages representation collapse by
penalizing rank-deficient centroid matrices.
Let $k:=\operatorname{rank}(\b{G})$ and
$r:=\min(b,p)$. 
As proved in supplementary
material \ref{app:csc_collapse_details}, for any feasible batch, we have:
\begin{equation}
\ell_{\mathrm{csc}}
\geq \frac{r-k}{r+\kappa k}.
\label{eq:csc_rank_gap_main}
\end{equation}
Consequently, for $r>1$, complete centroid
collapse incurs a strictly positive loss:
\begin{equation}
\ell_{\mathrm{csc}}
\geq \frac{r-1}{r+\kappa}>0.
\end{equation}
Thus, collapsed representations cannot be
global minimizers of CSC.

More generally, if $\ell_{\mathrm{csc}}
\leq\eta<1$, then (see proof in supplementary
material \ref{app:csc_collapse_details}):
\begin{equation}
\operatorname{rank}(\b{G})
\geq \frac{r(1-\eta)}{1+\kappa\eta}.
\end{equation}
Hence, sufficiently small loss guarantees
nontrivial representation diversity.

Furthermore, as shown in supplementary
material \ref{app:csc_collapse_details}, perfectly aligned constant-output
collapse is a strict saddle~\cite{ge2015escaping,boumal2023introduction} in the normalized
representation space, admitting directions
of decreasing loss. 
Detailed theoretical guarantees, including
centered-variance bounds and strict-saddle
analysis, are provided in the supplementary
material \ref{app:csc_collapse_details}.

\section{Experiments}
\label{sec:experiments}

We evaluate the effectiveness of Centroid Spectral Coupling
(CSC) in learning transferable visual representations.
Our primary objective is to assess whether the proposed
single-term spectral loss can achieve competitive
representation quality compared with established
non-contrastive self-supervised learning methods.

\subsection{Experimental Setup}
\label{subsec:experimental_setup}

\textbf{Datasets and evaluation protocol.}
We evaluate CSC on three image classification benchmarks:
CIFAR-10 and CIFAR-100~\cite{krizhevsky2009cifar},
and Imagenette~\cite{Howard_Imagenette_2019}.
All methods are evaluated using a ResNet-50
backbone~\cite{he2016resnet}.
Input crop resolutions are $32\times32$ for CIFAR-10
and CIFAR-100, and $128\times128$ for Imagenette.

Following standard self-supervised evaluation protocols,
we assess the quality of the learned representations
using linear probing and $k$-nearest-neighbor (kNN)
classification with frozen encoders.
Linear probing evaluates the linear separability
of the learned representations, while kNN classification
assesses their local discriminative structure.
We report Top-1 and Top-5 accuracy on held-out data.
For kNN evaluation, the number of neighbors is selected
based on validation Top-1 accuracy and subsequently
used for both Top-1 and Top-5 evaluation.

\textbf{Baselines.}
Our primary comparisons are with three established
non-contrastive methods: VICReg~\cite{bardes2022vicreg},
Barlow Twins~\cite{zbontar2021barlow}, and Maximum
Manifold Capacity Representations
(MMCR)~\cite{yerxa2023mmcr}.
These methods provide relevant comparisons because
they encourage representation diversity through
different regularization or geometric objectives.
In particular, MMCR provides a closely related
spectral baseline, optimizing a nuclear-norm
criterion over augmented-view centroids.

\begin{table*}[t]
\centering
\caption{
\textbf{Frozen ResNet-50 evaluation on CIFAR-10,
CIFAR-100, and Imagenette.}
Held-out classification accuracy (\%) using
linear probing and kNN classification (seed 0).
All evaluated methods are non-contrastive SSL approaches.
Input resolutions are $32\times32$ for CIFAR
and $128\times128$ for Imagenette.
\textbf{Bold} indicates the best result and
\underline{underlining} the second-best result
within each dataset and metric.
}
\label{tab:ssl_main_results}
\small
\begin{tabular*}{\textwidth}{
@{\extracolsep{\fill}}llcccc@{}
}
\toprule
& &
\multicolumn{2}{c}{Linear Probing} &
\multicolumn{2}{c}{kNN} \\
\cmidrule(lr){3-4}
\cmidrule(lr){5-6}
Dataset & Method & Top-1 & Top-5 & Top-1 & Top-5 \\
\midrule

\multirow{4}{*}{CIFAR-10}
& CSC
& \textbf{84.69}
& \underline{99.36}
& \textbf{85.31}
& \underline{98.56} \\

& MMCR (matched)
& 82.13
& 99.06
& \underline{82.73}
& 98.28 \\

& VICReg
& 81.97
& 99.09
& 79.62
& 98.05 \\

& Barlow Twins
& \underline{83.94}
& \textbf{99.42}
& 82.71
& \textbf{98.69} \\

\midrule

\multirow{4}{*}{CIFAR-100}
& CSC
& \textbf{56.92}
& \textbf{82.23}
& \textbf{55.21}
& \textbf{79.82} \\

& MMCR (matched)
& 54.50
& 80.61
& \underline{49.81}
& 75.31 \\

& VICReg
& 45.32
& 73.50
& 37.06
& 64.33 \\

& Barlow Twins
& \underline{54.77}
& \underline{81.17}
& 48.70
& \underline{75.34} \\

\midrule

\multirow{4}{*}{Imagenette}
& CSC
& \textbf{80.97}
& 96.79
& \underline{80.25}
& \underline{96.82} \\

& MMCR (matched)
& 78.90
& 97.02
& 79.39
& 96.48 \\

& VICReg
& \underline{79.95}
& \textbf{97.48}
& 79.41
& \textbf{97.17} \\

& Barlow Twins
& 78.27
& \underline{97.12}
& \textbf{80.64}
& 96.36 \\

\bottomrule
\end{tabular*}
\end{table*}

\subsection{Comparison with Non-Contrastive SSL Methods}
\label{subsec:noncontrastive_comparison}

Table~\ref{tab:ssl_main_results} summarizes the
performance of CSC and the competing SSL methods
across all three benchmarks.

\textbf{Linear evaluation.}
CSC consistently achieves the highest linear
Top-1 accuracy among the evaluated non-contrastive
methods across all three datasets.
Specifically, CSC obtains $84.69\%$ on CIFAR-10,
$56.92\%$ on CIFAR-100, and $80.97\%$ on Imagenette,
improving upon the strongest competing non-contrastive
baseline by $0.75$, $2.15$, and $1.02$ percentage
points, respectively.

The improvement is particularly notable on
CIFAR-100, where CSC outperforms the matched
MMCR baseline by $2.42$ percentage points
in linear Top-1 accuracy.
Since MMCR also optimizes a spectral criterion
based on augmented-view centroids, this comparison
is particularly relevant to the proposed method.
The results indicate that the ridge-spectral
formulation of CSC provides competitive
representations relative to an alternative
centroid-based spectral objective.

\textbf{Nearest-neighbor evaluation.}
CSC demonstrates particularly strong performance
under kNN evaluation on the CIFAR benchmarks.
On CIFAR-10, CSC achieves $85.31\%$ kNN
Top-1 accuracy, outperforming all evaluated
non-contrastive methods.
On CIFAR-100, the improvement becomes more
pronounced: CSC achieves $55.21\%$ kNN Top-1
accuracy, compared with $49.81\%$ for MMCR
and $48.70\%$ for Barlow Twins.
This corresponds to improvements of $5.40$
and $6.51$ percentage points, respectively.

CSC also achieves the highest kNN Top-5 accuracy
on CIFAR-100, reaching $79.82\%$.
These results suggest that the representations
learned by CSC exhibit useful local
discriminative structure, particularly on
CIFAR-100, where classification requires
distinguishing among a larger number of categories.

Unlike approaches that explicitly combine
multiple regularization objectives, CSC obtains
these results through a single spectral loss.
This observation is consistent with the
motivation underlying CSC: coupling
augmented-view alignment and representation
diversity within a unified objective.

\textbf{Performance on Imagenette.}
On Imagenette, CSC maintains the strongest
linear Top-1 performance among the evaluated
non-contrastive methods, achieving $80.97\%$.
However, its kNN Top-1 accuracy of $80.25\%$
is slightly below Barlow Twins at $80.64\%$.
Similarly, CSC does not attain the strongest
Top-5 accuracy on this dataset.

These results indicate that the advantages of CSC
vary across evaluation protocols and datasets.
While CSC provides consistent improvements
in linear Top-1 accuracy over the non-contrastive
baselines, its advantage in nearest-neighbor
classification is more pronounced on the
CIFAR benchmarks.

\subsection{Ablation Study on the Role of Centroid Magnitude}
\label{subsec:centroid_ablation}

To investigate the importance of preserving centroid
magnitudes, we conduct an ablation study comparing
three spectral self-supervised objectives:
the proposed CSC, CSC with centroid renormalization
(CSC + Renorm), and the matched MMCR baseline.
The CSC + Renorm variant applies $\ell_2$ normalization
to each centroid after averaging its two augmented
views, while keeping the ridge-spectral objective
unchanged. This modification removes the centroid
magnitude information that, in the original CSC,
encodes augmented-view agreement.

\begin{figure*}[t]
    \centering
    \includegraphics[width=0.48\textwidth]
    {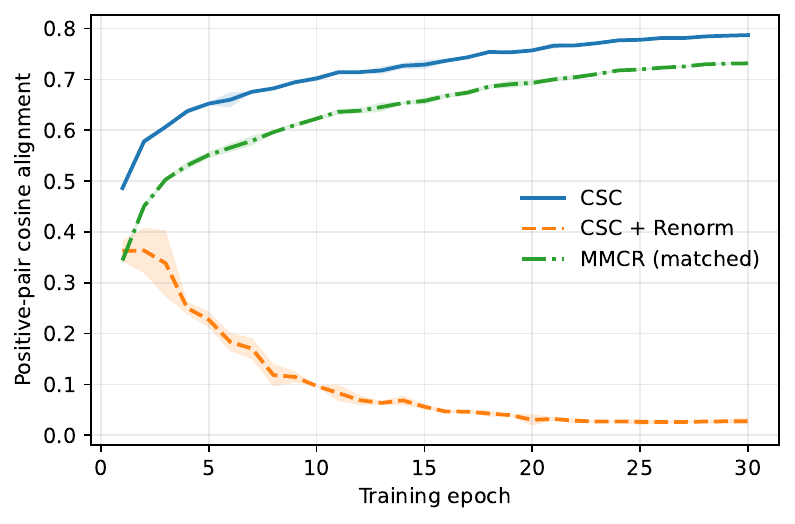}
    \hfill
    \includegraphics[width=0.48\textwidth]
    {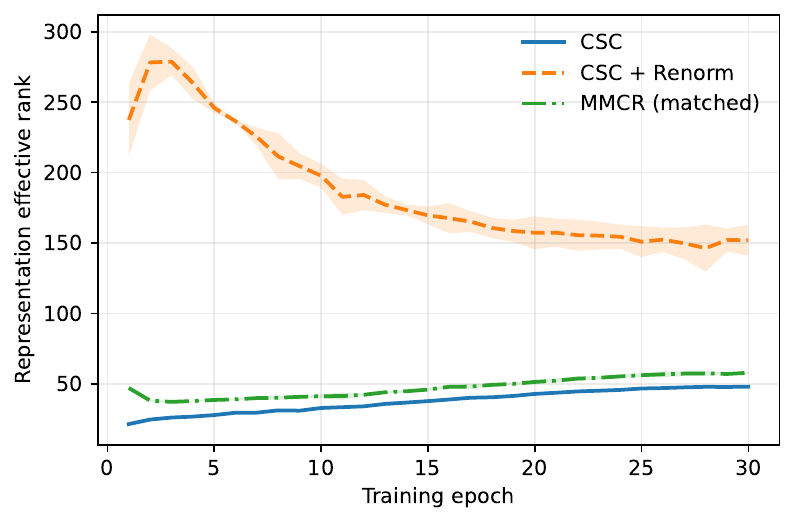}
    \caption{
    \textbf{Effect of centroid renormalization on
    augmented-view alignment and spectral diversity.}
    Training dynamics on CIFAR-10 using ResNet-50.
    \textbf{Left:} Positive-pair cosine alignment
    across training epochs.
    \textbf{Right:} Representation effective rank
    across training epochs.
    Unlike the original CSC, centroid renormalization
    substantially reduces positive-pair alignment
    despite producing a higher effective rank.
    CSC maintains strong alignment and nontrivial
    spectral diversity, illustrating the importance
    of preserving centroid magnitudes.
    }
    \label{fig:csc_centroid_ablation}
\end{figure*}

Figure~\ref{fig:csc_centroid_ablation} reveals a
pronounced difference in the representation dynamics
of the three objectives.
The original CSC progressively improves augmented-view
alignment, reaching a positive-pair cosine similarity
of approximately $0.79$ after 30 training epochs,
compared with approximately $0.73$ for MMCR.
In contrast, CSC + Renorm exhibits a substantial
deterioration in alignment, approaching zero
cosine similarity as training progresses.

Interestingly, the renormalized variant maintains
a considerably higher representation effective rank,
reaching approximately $150$, compared with
$48$ for CSC and $58$ for MMCR. Effective rank is also used in RankMe as a diagnostic for self-supervised representations ~\cite{garrido2023rankme}.
This observation highlights an important distinction:
promoting spectral diversity alone does not ensure
agreement between augmented views of the same instance.
Although centroid renormalization encourages a broader
distribution of representation energy, it removes
the magnitude-dependent alignment mechanism
present in the original CSC formulation.

These empirical observations are consistent with
our theoretical analysis in
Section~\ref{subsec:csc_within_instance},
which establishes the relationship between
centroid energy and augmented-view agreement.
The results provide evidence that preserving
centroid magnitudes is important for maintaining
alignment while encouraging representation diversity.

Furthermore, the comparison with MMCR reveals
different alignment--diversity trade-offs.
CSC achieves stronger positive-pair alignment,
whereas MMCR maintains a somewhat higher effective
rank. Thus, the results do not establish universal
superiority of either objective, but demonstrate
that the proposed ridge-spectral formulation
induces distinct representation dynamics.

\vspace{-0.6mm}

Overall, the ablation supports the central design
choice of CSC: retaining the magnitude information
of augmented-view centroids enables the spectral
objective to jointly encourage alignment and
representation diversity without introducing
a separate alignment penalty.

\subsection{Sensitivity to the Ridge Parameter}
\label{subsec:kappa_sensitivity}

We investigate how the ridge parameter $\kappa$
influences the alignment--diversity trade-off
underlying CSC. We evaluate
$\kappa\in\{1,10,100\}$ on CIFAR-100 using
ResNet-50, with identical training configurations
over 30 epochs and three random seeds.

\begin{figure}[t]
    \centering
    \includegraphics[width=0.86\linewidth]
    {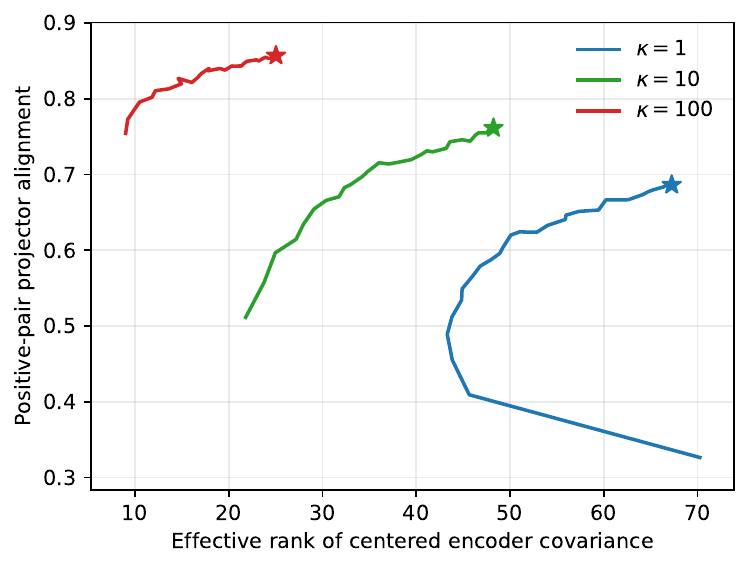}
    \caption{
    \textbf{Effect of the ridge parameter $\kappa$
    on alignment and representation diversity.}
    Training trajectories on CIFAR-100 using
    ResNet-50. Alignment is measured on normalized
    projector outputs, whereas effective rank is
    computed from centered encoder-feature covariance.
    Stars indicate the final training epoch.
    Increasing $\kappa$ favors alignment, while
    smaller values preserve greater spectral diversity.
    }
    \label{fig:kappa_sensitivity}
\end{figure}

Figure~\ref{fig:kappa_sensitivity} demonstrates
that $\kappa$ substantially influences the learned
representations. Increasing $\kappa$ from $1$ to
$100$ raises mean positive-pair alignment from
$0.710$ to $0.868$, while reducing effective rank
from $76.52$ to $24.18$.

An intermediate value, $\kappa=10$, achieves
the highest observed linear-probe accuracy
($47.67\%$), compared with $47.08\%$ and
$42.15\%$ for $\kappa=1$ and $\kappa=100$,
respectively.

These observations are consistent with our
theoretical characterization of the ridge
parameter: stronger emphasis on alignment
can reduce representation diversity, while
greater diversity alone does not guarantee
improved downstream performance.

The results highlight the importance of
balancing these complementary properties.
Additional numerical results and variability
across seeds are provided in the supplementary
material \ref{app:additional_experiments}.

\subsection{Discussion}
\label{subsec:experimental_discussion}

The experimental results highlight three
main observations.

First, CSC consistently delivers strong
linear evaluation performance among
non-contrastive SSL methods across the
three benchmarks. In particular, its
improvements over MMCR provide evidence
that the choice of spectral functional
can influence downstream representation
quality, even when both objectives operate
on augmented-view centroids.

Second, CSC achieves particularly strong
nearest-neighbor performance on CIFAR-100.
This suggests that the proposed objective
can produce representations with useful
local discriminative properties, beyond
their linear classification performance.


The centroid-renormalization ablation further
demonstrates that spectral diversity alone is
insufficient to maintain augmented-view agreement.
Preserving centroid magnitudes is therefore an
important design choice underlying the implicit
alignment mechanism of CSC.

\section{Conclusion}
\label{sec:conclusion}

In this paper, we introduced Centroid Spectral Coupling
(CSC), a self-supervised learning objective that
implicitly promotes augmented-view alignment,
representation diversity, and resistance to collapse
through a single ridge-spectral loss.
By applying a spectral functional to the uncentered
second-moment matrix of augmented-view centroids,
CSC couples these complementary properties without
requiring explicit contrastive negatives or separately
weighted regularization terms.

We provided a theoretical characterization of
this coupling, including an exact decomposition
of the objective into nonnegative alignment and
spectral-imbalance components. We further established
finite-loss guarantees for augmented-view agreement
and representation diversity, characterized the
geometry of global optima, and demonstrated that
complete constant-output collapse is a strict saddle
in the normalized representation space.

Experiments on CIFAR-10, CIFAR-100, and Imagenette
demonstrated competitive representation quality
relative to established self-supervised methods.
CSC achieved the highest linear Top-1 accuracy
among the evaluated non-contrastive baselines
across all three datasets and the strongest kNN Top-1 performance among the evaluated
non-contrastive methods on both CIFAR benchmarks.
These findings support the practical effectiveness
of the proposed unified spectral formulation.

Overall, CSC demonstrates that alignment,
representation diversity, and collapse resistance
can emerge from a single spectral objective,
providing a principled alternative to
multi-component non-contrastive SSL losses.
Future work will investigate its scalability
to larger datasets and architectures, as well as
the relationship between spectral geometry
and downstream semantic representation quality.

{
    \small
    \bibliographystyle{ieeenat_fullname}
    \bibliography{ref}
}

\clearpage

\appendix

{\huge
Supplementary Materials:
}

\section{Additional Theoretical Results}
\label{app:additional_theory}

\subsection{Implicit Representation Diversity}
\label{app:csc_diversity_details}

The second effect of CSC is to encourage a balanced
distribution of centroid energy across spectral directions.
Unlike the total centroid energy $t=\operatorname{tr}(\b{G})$,
which measures augmented-view agreement, the distribution
of eigenvalues provides information about representation
diversity across different instances.

We first establish a connection between the eigenvalues
of $\b{G}$ and the pairwise similarities among centroid
representations through the \textit{frame potential}.

\begin{proposition}[Frame potential and spectral balance]
\label{prop:csc_frame_potential}

Let $\b{C}=[\b{c}_1,\ldots,\b{c}_b]\in
\mathbb{R}^{p\times b}$ and $\b{G}$ be defined in Eq. (\ref{eq:csc_second_moment}).
Let $r=\min(b,p)$, and denote the $r$ potentially
nonzero eigenvalues of $\b{G}$, including zero padding
when necessary, by $\lambda_1,\ldots,\lambda_r$.
Let $t$ denote the total centroid energy defined in Eq. (\ref{equation_t_tr_G_sum_eig}).
Then:
\begin{equation}
\operatorname{tr}(\b{G}^2)
=
\sum_{j=1}^{r}\lambda_j^2
=
\frac{1}{b^2}
\sum_{i=1}^{b}\sum_{j=1}^{b}
(\b{c}_i^\top\b{c}_j)^2.
\label{eq:csc_frame_potential}
\end{equation}
Moreover:
\begin{equation}
\operatorname{tr}(\b{G}^2)
\ge\frac{t^2}{r},
\label{eq:csc_frame_lower}
\end{equation}
with equality if and only if:
\begin{equation}
\lambda_1=\cdots=\lambda_r=\frac{t}{r}.
\label{eq:csc_frame_equality}
\end{equation}
At fixed total energy $t$, this same flat spectrum
also minimizes the CSC loss.
More generally, we have:
\begin{equation}\label{equation_tr_G_squared_t2_r_sum_lambda_q}
\operatorname{tr}(\b{G}^2)
=
\frac{t^2}{r}
+
\sum_{j=1}^{r}(\lambda_j-q)^2,
\end{equation}
where $q=t/r$.

\end{proposition}

\textit{Proof.} See
supplementary material~\ref{app:csc_frame_potential_proof}.

The quantity $\operatorname{tr}(\b{G}^2)$ is the
normalized \textit{frame potential} of the centroid
representations. Equation~\eqref{eq:csc_frame_potential}
expresses this quantity in two equivalent forms.
From the spectral perspective, it measures the
concentration of energy among the eigenvalues of
$\b{G}$. From the geometric perspective, it measures
the aggregate squared inner products between centroid
vectors.

In particular, for unit-norm centroids,
$(\b{c}_i^\top\b{c}_j)^2$ in Eq. (\ref{eq:csc_frame_potential}) is zero for orthogonal
vectors and equals one for parallel or antiparallel
vectors. Thus, smaller aggregate squared inner
products indicate greater directional diversity.

Equation (\ref{equation_tr_G_squared_t2_r_sum_lambda_q}) shows that, at fixed $t$, the frame potential is
the sum of a constant and the squared deviations
of the eigenvalues from their mean. A flat
spectrum minimizes these deviations and hence
the frame potential.
This property is closely related to the spectral
imbalance penalty in the exact decomposition
of CSC. Both quantities are minimized at fixed
energy when all $r$ eigenvalues are equal.

We next characterize the geometric properties
of centroid representations at a global optimum
of CSC.

\begin{proposition}[Centroid geometry at zero CSC loss]
\label{prop:csc_zero_loss_geometry}

Suppose $\kappa>0$ and $\ell_{\mathrm{csc}}=0$.
Then, all centroids have unit norm and the
$r=\min(b,p)$ potentially nonzero eigenvalues
of $\b{G}$ satisfy:
\begin{equation}
\|\b{c}_i\|_2=1,
\qquad
\lambda_j=\frac{1}{r},
\end{equation}
for all $i=1,\ldots,b$ and $j=1,\ldots,r$.

Consequently:
\begin{equation}
\operatorname{tr}(\b{G}^2)=\frac{1}{r},
\qquad
\sum_{i\ne j}
(\b{c}_i^\top\b{c}_j)^2
=
\frac{b^2}{r}-b.
\label{eq:csc_optimal_frame}
\end{equation}

In particular, the following two cases arise.

\textit{Case 1: $b\le p$.}
In this case, $r=b$, and:
\begin{equation}
\b{c}_i^\top\b{c}_j=0,
\qquad i\ne j.
\label{eq:csc_optimal_orthogonality}
\end{equation}
Thus, all distinct centroids are mutually
orthogonal at zero CSC loss.

\textit{Case 2: $b>p$.}
In this case, $r=p$, and:
\begin{equation}
\b{G}
=
\frac{1}{b}\b{C}\b{C}^\top
=
\frac{1}{p}\b{I}_p.
\label{eq:csc_tight_frame}
\end{equation}
Together with the unit-norm condition,
this characterizes a \textit{unit-norm tight frame}.
\end{proposition}

\textit{Proof.} See
supplementary material~\ref{app:csc_zero_loss_geometry_proof}.

Proposition~\ref{prop:csc_zero_loss_geometry}
provides a geometric interpretation of the
spectral diversity encouraged by CSC.
When the embedding dimension is sufficiently
large ($p\ge b$), the globally optimal centroids
are mutually orthogonal, eliminating pairwise
directional correlations within the batch.
When $b>p$, mutual orthogonality of all centroids
is impossible. Instead, the optimal centroids
form a unit-norm tight frame, distributing their
second-moment energy equally across the
$p$ embedding directions.
Therefore, the global optima of CSC favor balanced spectral
energy, which encourages aggregate directional
diversity among sample centroids.

\subsection{Detailed Collapse-Resistance Guarantees}
\label{app:csc_collapse_details}

The preceding sections established that CSC encourages agreement
between augmented views of the same instance and a balanced
spectral distribution among instance centroids. We now show
that these two properties also give CSC a precise
\textit{collapse-resistance guarantee}: sufficiently small
CSC loss is incompatible with collapsed representations.
This distinction is important because augmented-view alignment
alone permits all instances to have the same representation.

We distinguish two forms of collapse. A \textit{rank-deficient}
representation has $\operatorname{rank}(\b{G})<r$, where
$r=\min(b,p)$. \textit{Complete centroid collapse} occurs when
$\b{c}_1=\cdots=\b{c}_b$. To characterize the latter, define
the mean centroid and centered covariance:
\begin{align}
&\b{\mu}:=\frac{1}{b}\sum_{i=1}^{b}\b{c}_i,
\label{eq:csc_centroid_mean} \\
&\b{\Sigma}
:=\frac{1}{b}\sum_{i=1}^{b}
(\b{c}_i-\b{\mu})(\b{c}_i-\b{\mu})^\top=\b{G}-\b{\mu}\b{\mu}^\top\succeq\b{0}.
\label{eq:csc_centered_cov}
\end{align}

\begin{lemma}[The pairwise-distance identity]\label{lemma_pairwise_distance_identity}
The trace of the centered covariance is the average squared deviation from the mean, which also admits the pairwise-distance expression:
\begin{equation}
\operatorname{tr}(\b{\Sigma})
=\frac{1}{2b^2}\sum_{i,j=1}^{b}
\|\b{c}_i-\b{c}_j\|_2^2.
\label{eq:csc_pairwise_diversity}
\end{equation}
Therefore, $\b{\Sigma}=\b{0}$ if and only if all centroids
are identical.
\end{lemma}
\begin{proof}
See supplementary material \ref{section_proof_lemma_pairwise_distance_identity}.
\end{proof}

\begin{proposition}[Loss gap for rank-deficient representations]
\label{prop:csc_finite_rank}
Let $\kappa>0$ and $k:=\operatorname{rank}(\b{G})$.
For every feasible batch, we have:
\begin{equation}
\ell_{\mathrm{csc}}\ge
\frac{r-k}{r+\kappa k}.
\label{eq:csc_rank_loss_gap}
\end{equation}
Moreover, if $\ell_{\mathrm{csc}}\le\eta<1$, with
$\eta\ge0$, then:
\begin{equation}
\operatorname{rank}(\b{G})\ge
\frac{r(1-\eta)}{1+\kappa\eta},\quad
\operatorname{rank}(\b{\Sigma})\ge
\operatorname{rank}(\b{G})-1.
\label{eq:csc_finite_rank}
\end{equation}
In particular, for $r>1$:
\begin{equation}
\ell_{\mathrm{csc}}<\frac{r-1}{r+\kappa}
\,\Longrightarrow\,
\b{\Sigma}\ne\b{0}.
\label{eq:csc_nocollapse_threshold}
\end{equation}
\end{proposition}

\textit{Proof.} See supplementary material~\ref{app:csc_finite_rank_proof}.

Proposition~\ref{prop:csc_finite_rank} gives a direct
mathematical explanation for collapse resistance. If all
centroids are identical, then $\operatorname{rank}(\b{\Sigma}) = 0$ so $\operatorname{rank}(\b{G})\le1$,
and the CSC loss cannot fall below
$(r-1)/(r+\kappa)>0$ for $r>1$.
For example, if $\b{u}_i=\b{v}_i=\b{w}$ for every
instance, with $\|\b{w}\|_2=1$, then
$\b{G}=\b{w}\b{w}^\top$ and the loss equals this
positive threshold. Consequently, such a collapsed
representation cannot be a global minimizer of CSC.

We next quantify how strongly the global optimum
excludes complete centroid collapse.

\begin{proposition}[Noncollapse of centroids at zero CSC loss]
\label{prop:csc_centered_nocollapse}
Suppose $r=\min(b,p)>1$ and $\ell_{\mathrm{csc}}=0$.
Then:
\begin{equation}
\operatorname{rank}(\b{\Sigma})\ge r-1,
\qquad
\operatorname{tr}(\b{\Sigma})\ge1-\frac{1}{r}.
\label{eq:csc_centered_bound}
\end{equation}
Consequently, the mean squared distance between
sample centroids satisfies:
\begin{equation}
\frac{1}{b^2}\sum_{i,j=1}^{b}
\|\b{c}_i-\b{c}_j\|_2^2
\ge2\left(1-\frac{1}{r}\right).
\label{eq:csc_pairwise_nocollapse}
\end{equation}
In particular, complete centroid collapse is impossible
at zero CSC loss.
\end{proposition}

\textit{Proof.} See
supplementary material~\ref{app:csc_centered_nocollapse_proof}.

The first bound in Eq. (\ref{eq:csc_centered_bound}) guarantees variation in at least $r-1$
centered directions. The second bound in Eq. (\ref{eq:csc_pairwise_nocollapse}) guarantees a strictly
positive amount of \textit{average inter-sample variation},
rather than merely requiring that some centroids differ.
These conclusions complement the geometric characterization
of zero-loss centroids in
Proposition~\ref{prop:csc_zero_loss_geometry}.

Finally, we establish a quantitative noncollapse guarantee
for losses that are small but not necessarily zero.

\begin{proposition}[Finite-loss centered-variance certificate]
\label{prop:csc_finite_variance}
For every feasible batch and $\kappa>0$, we have:
\begin{equation}
\operatorname{tr}(\b{\Sigma})\ge
\frac{\kappa}{1+\kappa}
\left[
\frac{r-1}{r+\kappa}-\ell_{\mathrm{csc}}
\right]_+,
\label{eq:csc_finite_variance}
\end{equation}
where $[\cdot]_+:=\max(\cdot, 0)$ is the positive-part operator. Therefore, for $r>1$,
any batch satisfying
$\ell_{\mathrm{csc}}<(r-1)/(r+\kappa)$ has
$\operatorname{tr}(\b{\Sigma})>0$, so its centroids
cannot all be identical.
\end{proposition}

\textit{Proof.} See supplementary material~\ref{app:csc_finite_variance_proof}.

\begin{proposition}[Instability of constant-output collapse]
\label{prop:csc_collapse_instability}
Let $b,p\ge2$ and $\kappa>0$. Consider the
completely collapsed configuration:
\begin{equation}
\b{u}_i=\b{v}_i=\b{w},
\qquad
\|\b{w}\|_2=1,
\qquad \forall i.
\end{equation}
Then:
\begin{equation}
\b{G}=\b{w}\b{w}^\top,
\qquad
\ell_{\mathrm{csc}}=\frac{r-1}{r+\kappa}.
\end{equation}
Although this configuration is stationary under
unit-norm constraints, it is not a local minimum
of CSC in the normalized representation space.
Specifically, there exist arbitrarily small
perturbations that preserve perfect augmented-view
alignment and unit norms, while strictly
decreasing the CSC loss by increasing
spectral diversity.
Consequently, the completely collapsed configuration
is a strict saddle of the CSC objective in the
normalized representation space.
\end{proposition}

\textit{Proof.} See
supplementary material~\ref{app:csc_collapse_instability_proof}.

Together, Propositions \ref{prop:csc_finite_rank}, \ref{prop:csc_centered_nocollapse}, \ref{prop:csc_finite_variance}, and \ref{prop:csc_collapse_instability} provide complementary
theoretical explanations for the collapse-resistance
properties of CSC. Propositions~\ref{prop:csc_finite_rank} and \ref{prop:csc_finite_variance} establish
that complete centroid collapse is excluded below
an explicit positive loss threshold, while
Proposition~\ref{prop:csc_centered_nocollapse} guarantees substantial representation
diversity at the global optimum. 

Importantly, Proposition~\ref{prop:csc_collapse_instability}
further shows that the collapse-resistance properties
of CSC are not restricted to the low-loss regime.
Specifically, a completely collapsed constant-output
configuration is not a local minimum in the normalized
representation space, even without assuming that
the CSC loss is below the noncollapse threshold.
This result identifies a spectral-diversity descent
direction away from perfectly aligned constant-output
collapse, complementing the finite-loss noncollapse
guarantees. However, the strict-saddle property in
the normalized representation space does not, by
itself, guarantee escape from collapsed configurations
during neural-network optimization.

In our empirical evaluations across different datasets,
we did not observe representation collapse, including
during the initial training epochs. These observations
suggest that CSC exhibits favorable collapse-resistance
properties in practical training, beyond the regime
covered by the theoretical guarantees. If collapse
occurs in an individual training run, restarting
optimization with a different random initialization
may provide a simple practical mitigation.

\section{Proofs}

\subsection{Proof of Proposition \ref{proposition_loss_csc_tr_G}}\label{section_proof_loss_csc_tr_G}

Since $\b{G}$ is symmetric and positive semidefinite, it
admits the eigendecomposition
$\b{G}=\b{Q}\b{\Lambda}\b{Q}^\top$, where $\b{Q}$ is
orthogonal and $\b{\Lambda}$ contains the eigenvalues of $\b{G}$. Therefore:
\begin{align*}
&\operatorname{tr}\left[
\b{G}(\b{G}+\tau\b{I}_p)^{-1}
\right] 
\\
&\overset{(a)}{=} \operatorname{tr}\left[
\b{Q}\b{\Lambda}\b{Q}^\top(\b{Q}\b{\Lambda}\b{Q}^\top+\tau\b{Q}\b{I}_p\b{Q}^\top)^{-1}
\right] 
\\
&\overset{(b)}{=} \operatorname{tr}\left[
\b{Q}\b{\Lambda}\b{Q}^\top(\b{Q}(\b{\Lambda}+\tau\b{I}_p)\b{Q}^\top)^{-1}
\right] 
\\
&\overset{(c)}{=} \operatorname{tr}\left[
\b{Q}\b{\Lambda}\b{Q}^\top((\b{Q}^{\top})^{-1}(\b{\Lambda}+\tau\b{I}_p)^{-1}\b{Q}^{-1})
\right] \\
&\overset{(d)}{=} \operatorname{tr}\left[
\b{Q}\b{\Lambda}\underbrace{\b{Q}^\top(\b{Q}}_{=\b{I}_p}(\b{\Lambda}+\tau\b{I}_p)^{-1}\b{Q}^\top)
\right] \\
&\overset{(e)}{=} \operatorname{tr}\left[
\b{Q}\b{\Lambda}
(\b{\Lambda}+\tau\b{I}_p)^{-1}
\b{Q}^\top
\right] \\
&\overset{(f)}{=} \operatorname{tr}\left[
\b{\Lambda}
(\b{\Lambda}+\tau\b{I}_p)^{-1}
\underbrace{\b{Q}^\top \b{Q}}_{=\b{I}_p}
\right] 
\end{align*}
\begin{align*}
&\overset{(g)}{=}\operatorname{tr}\left[
\b{\Lambda}(\b{\Lambda}+\tau\b{I}_p)^{-1}
\right]
\overset{(h)}{=}\sum_{j=1}^{d}\frac{\lambda_j}{\lambda_j+\tau}
\overset{(i)}{=}\sum_{j=1}^{r}\frac{\lambda_j}{\lambda_j+\kappa/r},
\end{align*}
where the step $(a)$ is because of substitution of eigendecomposition
$\b{G}=\b{Q}\b{\Lambda}\b{Q}^\top$ and the fact of $\b{Q}\b{\Lambda}\b{Q}^\top$ since $\b{Q}$ is a non-truncated orthogonal matrix. The step $(b)$ is because of factoring out the terms, the step $(c)$ is because inverse of multiplication of matrices reverses the order of matrix multiplications, the step $(d)$ is because the matrix $\b{Q}$ is orthogonal so $\b{Q}^\top = \b{Q}^{-1}$, and the steps $(e)$ and $(g)$ are because the matrix $\b{Q}$ is orthogonal so $\b{Q}^\top \b{Q} = \b{I}_p$ which does not have effect in matrix multiplication. The step $(f)$ is because of the cyclic property of trace, the step $(h)$ is because of the scalar format of the trace of matrix, and the step $(i)$ is because $\operatorname{rank}(\b{G})\leq r=\min(b,p)$, and hence the remaining eigenvalues are zero.

\subsection{Proof of Proposition \ref{prop:csc_optimum}}\label{section_proof_csc_optimum}

Recall Eqs.~(\ref{eq:csc_f_tau}) and
(\ref{eq:csc_concavity}). According to the first
and second derivatives $f_\tau'(\lambda)$ and
$f_\tau''(\lambda)$, the function $f_\tau$ is
strictly increasing and strictly concave on
$[0,\infty)$.

Since $f_\tau$ is concave, Jensen's inequality gives:
\begin{equation}
\frac{1}{r}
\sum_{j=1}^{r}f_\tau(\lambda_j)
\le
f_\tau\left(
\frac{1}{r}\sum_{j=1}^{r}\lambda_j
\right).
\label{eq:csc_alignment_proof_jensen}
\end{equation}
Using $\sum_{j=1}^{r}\lambda_j=t$ from
Eq.~(\ref{equation_t_tr_G_sum_eig}), we obtain:
\begin{align}
1-\ell_{\mathrm{csc}}
&\overset{(\ref{eq:csc_loss_f})}{=}
\frac{1+\kappa}{r}
\sum_{j=1}^{r}f_\tau(\lambda_j)
\nonumber\\
&\overset{(\ref{eq:csc_alignment_proof_jensen})}{\le}
(1+\kappa)
f_\tau\left(\frac{t}{r}\right)
\nonumber\\
&\overset{(\ref{eq:csc_f_tau})}{=}
(1+\kappa)
\frac{t/r}{t/r+\tau}
\nonumber\\
&\overset{(\ref{equation_tau})}{=}
(1+\kappa)
\frac{t/r}{t/r+\kappa/r}
\nonumber\\
&=
\frac{(1+\kappa)t}{t+\kappa}.
\label{eq:csc_alignment_proof_score_bound}
\end{align}

Since $0\le t\le1$ and $\kappa>0$, we have:
\begin{align*}
\frac{(1+\kappa)t}{t+\kappa}
&\le1
\nonumber\\
&\Longleftrightarrow
(1+\kappa)t\le t+\kappa
\nonumber\\
&\Longleftrightarrow
\kappa t\le\kappa
\nonumber\\
&\Longleftrightarrow
t\le1.
\end{align*}
Therefore:
\begin{equation*}
1-\ell_{\mathrm{csc}}
\le
\frac{(1+\kappa)t}{t+\kappa}
\le1.
\end{equation*}

Moreover, since $\lambda_j\ge0$ and
$\tau=\kappa/r>0$, we have:
\begin{equation*}
f_\tau(\lambda_j)
=
\frac{\lambda_j}{\lambda_j+\tau}
\ge0.
\end{equation*}
Consequently:
\begin{equation*}
1-\ell_{\mathrm{csc}}
=
\frac{1+\kappa}{r}
\sum_{j=1}^{r}f_\tau(\lambda_j)
\ge0.
\end{equation*}
Combining these results yields:
\begin{equation*}
0\le
1-\ell_{\mathrm{csc}}
\le
\frac{(1+\kappa)t}{t+\kappa}
\le1.
\end{equation*}
Subtracting the inequalities
$0\le1-\ell_{\mathrm{csc}}\le1$
from one gives:
\begin{equation*}
0\le\ell_{\mathrm{csc}}\le1.
\end{equation*}

We now characterize the conditions under which
the minimum loss $\ell_{\mathrm{csc}}=0$
is attained.

\textit{Necessity.}
Suppose that $\ell_{\mathrm{csc}}=0$.
Then $1-\ell_{\mathrm{csc}}=1$, so:
\begin{equation*}
1
\le
\frac{(1+\kappa)t}{t+\kappa}
\le1.
\end{equation*}
Hence, equality must hold:
\begin{equation*}
\frac{(1+\kappa)t}{t+\kappa}=1.
\end{equation*}
Multiplying both sides by $t+\kappa>0$:
\begin{align}
(1+\kappa)t
=t+\kappa
&\implies 
t+\kappa t
=t+\kappa
\nonumber\\
&\implies \kappa t
=\kappa.
\end{align}
Since $\kappa>0$, we obtain:
\begin{equation}
t=1.
\label{eq:csc_optimum_trace_one}
\end{equation}
Recall that:
\begin{equation}\label{equation_t_ib_sum_norm_c_2}
t=
\frac{1}{b}
\sum_{i=1}^{b}\|\b{c}_i\|_2^2,
\end{equation}
where:
\begin{equation*}
\b{c}_i
=
\frac{\b{u}_i+\b{v}_i}{2},
\qquad
\|\b{u}_i\|_2,\|\b{v}_i\|_2\le1.
\end{equation*}
By the triangle inequality:
\begin{align}\label{equation_c_norm_half_u_v__u_v_2_1}
\|\b{c}_i\|_2
&=
\frac12\|\b{u}_i+\b{v}_i\|_2
\le
\frac{
\|\b{u}_i\|_2+\|\b{v}_i\|_2
}{2}
\le1.
\end{align}
Thus:
\begin{equation}\label{equation_0_norm_c2_1}
0\le\|\b{c}_i\|_2^2\le1.
\end{equation}
According to Eqs. (\ref{eq:csc_optimum_trace_one}) and (\ref{equation_t_ib_sum_norm_c_2}), the average of $\|\b{c}_i\|_2^2$ equals $t=1$, so according to Eq. (\ref{equation_0_norm_c2_1}), every term must equal one:
\begin{equation}
\|\b{c}_i\|_2=1,
\qquad i=1,\ldots,b.
\label{eq:csc_optimum_unit_centroids}
\end{equation}

We next show that this condition forces
the two augmented views to be identical
unit vectors.
Using the parallelogram identity:
\begin{equation*}
\|\b{u}_i+\b{v}_i\|_2^2
+
\|\b{u}_i-\b{v}_i\|_2^2
=
2\|\b{u}_i\|_2^2+
2\|\b{v}_i\|_2^2.
\end{equation*}
Since, according to Eq. (\ref{equation_c_norm_half_u_v__u_v_2_1}):
\begin{equation}
\|\b{u}_i+\b{v}_i\|_2^2
=
4\|\b{c}_i\|_2^2=4,
\end{equation}
we obtain:
\begin{align*}
&4+\|\b{u}_i-\b{v}_i\|_2^2
=
2\|\b{u}_i\|_2^2+
2\|\b{v}_i\|_2^2
\le4.
\\
&\implies 
\|\b{u}_i-\b{v}_i\|_2^2\le0.
\end{align*}
Since squared Euclidean distances are
nonnegative, this implies:
\begin{equation*}
\|\b{u}_i-\b{v}_i\|_2^2=0.
\implies 
\b{u}_i=\b{v}_i.
\end{equation*}
Combining this with
$\|\b{c}_i\|_2=1$ gives:
\begin{equation}
\b{u}_i=\b{v}_i=\b{c}_i,
\qquad
\|\b{u}_i\|_2
=
\|\b{v}_i\|_2=1.
\label{eq:csc_optimum_identical_views}
\end{equation}
Thus, zero CSC loss necessarily implies
perfect agreement between the two
unit-normalized views of every instance.

It remains to establish the required
distribution of the eigenvalues.
Recall that Jensen's inequality gave:
\begin{equation*}
\frac1r\sum_{j=1}^{r}f_\tau(\lambda_j)
\le
f_\tau\left(\frac{t}{r}\right).
\end{equation*}
Since $\ell_{\mathrm{csc}}=0$ and $t=1$,
the two sides must be equal:
\begin{equation*}
\frac1r\sum_{j=1}^{r}f_\tau(\lambda_j)
=
f_\tau\left(\frac1r\right).
\end{equation*}
Because $f_\tau$ is strictly concave,
equality in Jensen's inequality holds
if and only if all its arguments are equal.
Therefore:
\begin{equation*}
\lambda_1=\lambda_2=\cdots=\lambda_r.
\end{equation*}
Since their sum is $t=1$:
\begin{equation*}
\sum_{j=1}^{r}\lambda_j
=
r\lambda_1=1.
\end{equation*}
Consequently:
\begin{equation*}
\lambda_1=\cdots=\lambda_r=\frac1r.
\label{eq:csc_optimum_flat_spectrum}
\end{equation*}
We have thus proved that zero CSC loss
requires both identical unit-norm
augmented views and a flat centroid
spectrum with eigenvalues $1/r$.

\textit{Sufficiency.}
Conversely, suppose that:
\begin{equation*}
\b{u}_i=\b{v}_i,
\qquad
\|\b{u}_i\|_2
=
\|\b{v}_i\|_2=1,
\end{equation*}
for every $i$, and:
\begin{equation*}
\lambda_j=\frac1r,
\qquad j=1,\ldots,r.
\end{equation*}
Substituting these eigenvalues into
the CSC objective yields:
\begin{align*}
\ell_{\mathrm{csc}}
&=
1-\frac{1+\kappa}{r}
\sum_{j=1}^{r}
\frac{\lambda_j}{\lambda_j+\kappa/r}
\nonumber\\
&=
1-\frac{1+\kappa}{r}
\sum_{j=1}^{r}
\frac{1/r}{1/r+\kappa/r}
\nonumber\\
&=
1-\frac{1+\kappa}{r}
\sum_{j=1}^{r}\frac1{1+\kappa}
\nonumber\\
&=
1-\frac{1+\kappa}{r}
\frac{r}{1+\kappa}
\nonumber\\
&=0.
\end{align*}
Therefore, the stated conditions
are also sufficient for achieving
zero CSC loss.

Such configurations are attainable
in the unconstrained representation space:
when $b\le p$, one may choose $b$
orthonormal unit centroids; when $b>p$,
one may choose a unit-norm tight frame
of $b$ vectors in $\mathbb{R}^p$.

Thus, the global minimum of the CSC
objective over feasible normalized
representations is zero, attained
exactly when the augmented-view pairs
are identical unit vectors and all
$r$ potentially nonzero eigenvalues
of $\b{G}$ equal $1/r$.

\subsection{Effect of Centroid Renormalization}
\label{app:csc_centroid_renormalization}

We provide a counterexample demonstrating why the
centroids in CSC should not be renormalized after
averaging the two augmented views. In particular,
we show that centroid renormalization can produce
zero CSC loss even when the corresponding augmented
views are substantially misaligned.

Consider a batch with $b=p\ge2$, and let
$\{\b{e}_1,\ldots,\b{e}_p\}$ be the standard
orthonormal basis of $\mathbb{R}^p$, with cyclic
indexing $\b{e}_{p+1}=\b{e}_1$.
For any $s\in(0,1]$, define the two augmented-view
representations of instance $i$ as:
\begin{align*}
\b{u}_i
&=
\sqrt{s}\,\b{e}_i
+\sqrt{1-s}\,\b{e}_{i+1},\\
\b{v}_i
&=
\sqrt{s}\,\b{e}_i
-\sqrt{1-s}\,\b{e}_{i+1}.
\end{align*}
Since $\b{e}_i^\top\b{e}_{i+1}=0$ and
$\|\b{e}_i\|_2=\|\b{e}_{i+1}\|_2=1$,
both representations have unit norm:
\begin{align*}
\|\b{u}_i\|_2^2
&=s+(1-s)=1,\\
\|\b{v}_i\|_2^2
&=s+(1-s)=1.
\end{align*}
However, their inner product is:
\begin{align*}
\b{u}_i^\top\b{v}_i
&=
s-(1-s)\\
&=2s-1.
\end{align*}
Thus, their cosine similarity approaches $-1$
as $s\to0^+$, meaning that the two augmented
views become nearly opposite.

We first evaluate the original CSC objective,
which does not renormalize centroids.
By definition:
\begin{align*}
\b{c}_i
&=\frac{\b{u}_i+\b{v}_i}{2}=\sqrt{s}\,\b{e}_i.
\end{align*}
Consequently:
\begin{equation*}
\|\b{c}_i\|_2^2=s.
\end{equation*}
The corresponding uncentered second-moment
matrix is:
\begin{align*}
\b{G}
&=\frac{1}{b}\sum_{i=1}^{b}
\b{c}_i\b{c}_i^\top=\frac{s}{p}\sum_{i=1}^{p}
\b{e}_i\b{e}_i^\top=\frac{s}{p}\b{I}_p,
\end{align*}
where we used $b=p$ and the completeness
relation for an orthonormal basis:
\begin{equation*}
\sum_{i=1}^{p}\b{e}_i\b{e}_i^\top
=\b{I}_p.
\end{equation*}
Therefore, all $r=p$ eigenvalues of
$\b{G}$ are equal:
\begin{equation*}
\lambda_1=\cdots=\lambda_r=\frac{s}{r}.
\end{equation*}
Substituting these eigenvalues into the
original CSC objective gives:
\begin{align*}
\ell_{\mathrm{csc}}
&=
1-\frac{1+\kappa}{r}
\sum_{j=1}^{r}
\frac{\lambda_j}{\lambda_j+\kappa/r}\\
&=
1-\frac{1+\kappa}{r}
\sum_{j=1}^{r}
\frac{s/r}{s/r+\kappa/r}\\
&=
1-\frac{(1+\kappa)s}{s+\kappa}\\
&=
\frac{\kappa(1-s)}{s+\kappa}.
\end{align*}
Since $\kappa>0$, this loss is strictly
positive whenever $s<1$, and it approaches
one as $s\to0^+$:
\begin{equation*}
\lim_{s\to0^+}\ell_{\mathrm{csc}}=1.
\end{equation*}
Thus, the original CSC objective penalizes
the disagreement between augmented views,
even though the centroid spectrum is
perfectly balanced for every $s>0$.

Now, consider a modified CSC objective in
which each centroid is renormalized to unit
norm after averaging:
\begin{equation*}
\hat{\b{c}}_i
:=
\frac{\b{c}_i}{\|\b{c}_i\|_2}.
\end{equation*}
Since $s>0$, the centroids are nonzero,
and their normalized versions are:
\begin{align*}
\hat{\b{c}}_i
&=
\frac{\sqrt{s}\,\b{e}_i}{\sqrt{s}}=\b{e}_i.
\end{align*}
The modified second-moment matrix becomes:
\begin{align*}
\hat{\b{G}}
&=
\frac{1}{b}\sum_{i=1}^{b}
\hat{\b{c}}_i\hat{\b{c}}_i^\top=
\frac{1}{p}\sum_{i=1}^{p}
\b{e}_i\b{e}_i^\top=\frac{1}{p}\b{I}_p.
\end{align*}
Therefore, its $r=p$ eigenvalues are:
\begin{equation*}
\hat{\lambda}_1=\cdots
=\hat{\lambda}_r=\frac{1}{r}.
\end{equation*}
The modified CSC loss is consequently:
\begin{align*}
\hat{\ell}_{\mathrm{csc}}
&=
1-\frac{1+\kappa}{r}
\sum_{j=1}^{r}
\frac{\hat{\lambda}_j}
{\hat{\lambda}_j+\kappa/r}\\
&=
1-\frac{1+\kappa}{r}
\sum_{j=1}^{r}
\frac{1/r}{1/r+\kappa/r}\\
&=
1-\frac{1+\kappa}{r}
\sum_{j=1}^{r}\frac{1}{1+\kappa}=0.
\end{align*}
Remarkably, this modified loss equals zero
for every $s\in(0,1]$, regardless of the
agreement between the augmented views.

To quantify their disagreement, observe that:
\begin{align*}
&\b{u}_i-\b{v}_i
=2\sqrt{1-s}\,\b{e}_{i+1},\\
&\|\b{u}_i-\b{v}_i\|_2^2
=4(1-s).
\end{align*}
Summing over all $b$ instances gives:
\begin{equation*}
\|\b{U}-\b{V}\|_F^2=4b(1-s),
\end{equation*}
and hence:
\begin{equation*}
\frac{1}{b}\|\b{U}-\b{V}\|_F^2
=4(1-s).
\end{equation*}
As $s\to0^+$, the average squared
augmented-view distance approaches its
maximum possible value of four:
\begin{equation*}
\lim_{s\to0^+}
\frac{1}{b}\|\b{U}-\b{V}\|_F^2=4,
\end{equation*}
while the modified loss remains zero:
\begin{equation*}
\hat{\ell}_{\mathrm{csc}}=0.
\end{equation*}
In contrast, the original CSC loss
approaches one in the same limit.

This counterexample demonstrates that
renormalizing the centroids removes the
magnitude information encoding agreement
between augmented views. Consequently,
the modified objective can achieve its
global minimum despite severe
augmented-view disagreement.

Therefore, retaining the original
unnormalized centroids is essential
for the intended coupling between
augmented-view alignment and spectral
diversity in CSC.

\subsection{Proof of the Ridge-Parameter Limits}
\label{app:csc_kappa_limits}
Recall that the CSC score is defined as:
\begin{equation*}
s_\kappa
:=1-\ell_{\mathrm{csc}}
=\frac{1+\kappa}{r}
\sum_{j=1}^{r}
\frac{\lambda_j}{\lambda_j+\kappa/r},
\end{equation*}
where $\kappa>0$, $r=\min(b,p)$, and
$\lambda_1,\ldots,\lambda_r\ge0$ are the
potentially nonzero eigenvalues of $\b{G}$.
Throughout this proof, $\b{G}$ and its
eigenvalues are held fixed as $\kappa$ varies.

We first consider the limit as
$\kappa\to0^+$. For each eigenvalue
$\lambda_j$, we have:
\begin{equation*}
\lim_{\kappa\to0^+}
\frac{\lambda_j}{\lambda_j+\kappa/r}
=
\begin{cases}
1, & \lambda_j>0,\\
0, & \lambda_j=0.
\end{cases}
\end{equation*}
Indeed, when $\lambda_j>0$, the
denominator converges to $\lambda_j$.
When $\lambda_j=0$, the fraction equals
zero for every $\kappa>0$.

Since the sum contains finitely many
terms, the limit can be taken term by
term. Moreover, $(1+\kappa)/r$
converges to $1/r$. Thus:
\begin{align*}
\lim_{\kappa\to0^+}s_\kappa
&=
\frac{1}{r}
\sum_{j=1}^{r}
\lim_{\kappa\to0^+}
\frac{\lambda_j}{\lambda_j+\kappa/r}\\
&=
\frac{1}{r}
\sum_{j=1}^{r}
\mathbb{1}_{\{\lambda_j>0\}}\\
&=
\frac{\operatorname{rank}(\b{G})}{r}.
\end{align*}
The last equality follows because
$\b{G}$ is positive semidefinite,
so its rank equals the number of
strictly positive eigenvalues.

We next consider the limit as
$\kappa\to\infty$. Multiplying the
numerator and denominator of each
spectral fraction by $r$ gives:
\begin{align*}
s_\kappa
&=
(1+\kappa)
\sum_{j=1}^{r}
\frac{\lambda_j}{r\lambda_j+\kappa}\\
&=
\frac{1+\kappa}{\kappa}
\sum_{j=1}^{r}
\frac{\lambda_j}{1+r\lambda_j/\kappa}.
\end{align*}
Since the eigenvalues remain fixed,
we have:
\begin{equation*}
\lim_{\kappa\to\infty}
\frac{1+\kappa}{\kappa}=1,
\end{equation*}
and, for every $j$:
\begin{equation*}
\lim_{\kappa\to\infty}
\frac{\lambda_j}{1+r\lambda_j/\kappa}
=\lambda_j.
\end{equation*}
Consequently:
\begin{align*}
\lim_{\kappa\to\infty}s_\kappa
&=
\sum_{j=1}^{r}\lambda_j\\
&=
\operatorname{tr}(\b{G})\\
&=t.
\end{align*}
Therefore, the two limiting expressions are:
\begin{equation*}
\lim_{\kappa\to0^+}s_\kappa
=\frac{\operatorname{rank}(\b{G})}{r},
\qquad
\lim_{\kappa\to\infty}s_\kappa=t.
\end{equation*}
This completes the proof.

\subsection{Proof of Proposition \ref{prop:csc_energy_alignment}}\label{section_proof_csc_energy_alignment}

We first investigate the squared norm of each centroid.
By its definition:
\begin{align}
\|\b{c}_i\|_2^2
&\overset{(\ref{eq:csc_centroid})}{=}
\left\|
\frac{\b{u}_i+\b{v}_i}{2}
\right\|_2^2
\nonumber\\
&=
\frac{1}{4}
(\b{u}_i+\b{v}_i)^\top
(\b{u}_i+\b{v}_i)
\nonumber\\
&=
\frac{1}{4}
\left(
\|\b{u}_i\|_2^2+
\|\b{v}_i\|_2^2+
2\b{u}_i^\top\b{v}_i
\right).
\label{eq:csc_centroid_norm_proof}
\end{align}
Since both views have unit norm, we obtain:
\begin{align}
\|\b{c}_i\|_2^2
&=
\frac{1}{4}
\left(2+2\b{u}_i^\top\b{v}_i\right)
=
\frac{1}{2}
\left(1+\b{u}_i^\top\b{v}_i\right).
\label{eq:csc_centroid_cosine_proof}
\end{align}
Because the representations are normalized,
$\b{u}_i^\top\b{v}_i$ equals their cosine similarity.
Thus:
\begin{equation}
\|\b{c}_i\|_2^2
=
\frac{1+\cos(\b{u}_i,\b{v}_i)}{2}.
\end{equation}

Substituting Eq.~\eqref{eq:csc_centroid_cosine_proof}
into Eq.~\eqref{eq:csc_trace_centroid_norm} gives:
\begin{align}
t
&=
\frac{1}{2b}
\sum_{i=1}^{b}
\left(1+\b{u}_i^\top\b{v}_i\right)
=
\frac{1}{2}
+\frac{1}{2b}
\sum_{i=1}^{b}\b{u}_i^\top\b{v}_i.
\label{eq:csc_trace_cosine_proof}
\end{align}

We now express this result in terms of the squared
distance between the two augmented views.
Since $\|\b{u}_i\|_2=\|\b{v}_i\|_2=1$, we have:
\begin{align*}
\|\b{u}_i-\b{v}_i\|_2^2
&=
\|\b{u}_i\|_2^2+
\|\b{v}_i\|_2^2-
2\b{u}_i^\top\b{v}_i
\nonumber\\
&=
2-2\b{u}_i^\top\b{v}_i.
\end{align*}
Rearranging gives:
\begin{equation}
\b{u}_i^\top\b{v}_i
=
1-\frac{1}{2}
\|\b{u}_i-\b{v}_i\|_2^2.
\label{eq:csc_inner_distance}
\end{equation}
Substituting Eq.~\eqref{eq:csc_inner_distance}
into Eq.~\eqref{eq:csc_trace_cosine_proof}:
\begin{align}
t
&=
\frac{1}{2}
+\frac{1}{2b}
\sum_{i=1}^{b}
\left(
1-\frac{1}{2}
\|\b{u}_i-\b{v}_i\|_2^2
\right)
\nonumber\\
&=
\frac{1}{2}
+\frac{1}{2}
-\frac{1}{4b}
\sum_{i=1}^{b}
\|\b{u}_i-\b{v}_i\|_2^2
\nonumber\\
&=
1-\frac{1}{4b}
\|\b{U}-\b{V}\|_F^2.
\label{eq:csc_trace_distance_proof}
\end{align}
This proves the first claim.

To prove the second claim, recall that:
\begin{align*}
\b{c}_i=\frac{\b{u}_i+\b{v}_i}{2} \\
\implies \b{u}_i-\b{c}_i
&=
\b{u}_i-
\frac{\b{u}_i+\b{v}_i}{2}
=
\frac{\b{u}_i-\b{v}_i}{2},
\end{align*}
and similarly:
\begin{equation*}
\b{v}_i-\b{c}_i
=
\frac{\b{v}_i-\b{u}_i}{2}.
\end{equation*}
Consequently:
\begin{equation}
\|\b{u}_i-\b{c}_i\|_2^2
=
\|\b{v}_i-\b{c}_i\|_2^2
=
\frac{1}{4}
\|\b{u}_i-\b{v}_i\|_2^2.
\end{equation}
Substituting into the definition of
$w_{\mathrm{pair}}$, i.e., Eq. (\ref{eq:csc_pair_scatter}), gives:
\begin{align}
w_{\mathrm{pair}}
&=
\frac{1}{2b}
\sum_{i=1}^{b}
\left(
\frac{1}{4}\|\b{u}_i-\b{v}_i\|_2^2
+
\frac{1}{4}\|\b{u}_i-\b{v}_i\|_2^2
\right)
\nonumber\\
&=
\frac{1}{4b}
\sum_{i=1}^{b}
\|\b{u}_i-\b{v}_i\|_2^2
=
\frac{1}{4b}
\|\b{U}-\b{V}\|_F^2.
\end{align}
Combining this expression with
Eq.~\eqref{eq:csc_trace_distance_proof} yields:
\begin{equation}
w_{\mathrm{pair}}=1-t.
\end{equation}
Thus, maximizing $t$ is equivalent to minimizing
the within-instance scatter $w_{\mathrm{pair}}$.

\subsection{Proof of Proposition \ref{prop:csc_decomposition}}\label{section_proof_csv_decomposition}

Recall Eq. (\ref{eq:csc_loss_f}) for the CSC loss function:
\begin{equation}
\ell_{\mathrm{csc}}
=
1-\frac{1+\kappa}{r}
\sum_{j=1}^{r}f_\tau(\lambda_j),
\label{eq:csc_loss_proof}
\end{equation}
where:
\begin{equation}\label{equation_f_lambda_on_lambda_plus_tau}
f_\tau(\lambda)
=
\frac{\lambda}{\lambda+\tau}
=
1-\frac{\tau}{\lambda+\tau},
\qquad
\tau=\frac{\kappa}{r}>0.
\end{equation}
Furthermore, let:
\begin{equation}
t=\sum_{j=1}^{r}\lambda_j,
\qquad
q=\frac{t}{r}.
\label{eq:csc_mean_eigenvalue}
\end{equation}
Thus, $q$ is the mean of the $r$ eigenvalues. Consequently:
\begin{align}
\sum_{j=1}^{r}(\lambda_j-q)
&=\sum_{j=1}^{r}\lambda_j-rq =t-r\frac{t}{r}=0.
\label{eq:csc_zero_deviation}
\end{align}

We first establish an exact algebraic identity. For $a>0$ and
$a+h>0$, we have:
\begin{equation}
\frac{1}{a+h}
=
\frac{1}{a}-\frac{h}{a(a+h)}.
\label{eq:csc_reciprocal_first}
\end{equation}
Indeed, the right-hand side simplifies to:
\begin{equation*}
\frac{1}{a}-\frac{h}{a(a+h)}
=
\frac{a+h-h}{a(a+h)}
=
\frac{1}{a+h}.
\end{equation*}
Applying Eq.~\eqref{eq:csc_reciprocal_first} once more to
the reciprocal appearing in its right-hand side gives:
\begin{align}
\frac{1}{a+h}
&=
\frac{1}{a}
-\frac{h}{a}\frac{1}{a+h}
\nonumber\\
&=
\frac{1}{a}
-\frac{h}{a}
\left(
\frac{1}{a}-\frac{h}{a(a+h)}
\right)
\nonumber\\
&=
\frac{1}{a}
-\frac{h}{a^2}
+\frac{h^2}{a^2(a+h)}.
\label{eq:csc_reciprocal_general}
\end{align}
Now, setting:
\begin{equation*}
a=q+\tau,
\qquad
h=\lambda-q,
\end{equation*}
we obtain $a+h=\lambda+\tau$, and therefore:
\begin{equation}
\frac{1}{\lambda+\tau}
=
\frac{1}{q+\tau}
-\frac{\lambda-q}{(q+\tau)^2}
+\frac{(\lambda-q)^2}
{(q+\tau)^2(\lambda+\tau)}.
\label{eq:csc_reciprocal_identity}
\end{equation}
Substituting Eq.~\eqref{eq:csc_reciprocal_identity}
into Eq. (\ref{equation_f_lambda_on_lambda_plus_tau}) yields:
\begin{align}
f_\tau(\lambda)
&=
1-\tau
\left[
\frac{1}{q+\tau}
-\frac{\lambda-q}{(q+\tau)^2}
+\frac{(\lambda-q)^2}
{(q+\tau)^2(\lambda+\tau)}
\right]
\nonumber\\
&=
1-\frac{\tau}{q+\tau}
+\frac{\tau(\lambda-q)}{(q+\tau)^2}
-\frac{\tau(\lambda-q)^2}
{(q+\tau)^2(\lambda+\tau)}.
\label{eq:csc_function_expansion}
\end{align}
Since:
\begin{equation*}
1-\frac{\tau}{q+\tau}
=
\frac{q}{q+\tau}
=
f_\tau(q),
\end{equation*}
we can rewrite Eq.~\eqref{eq:csc_function_expansion} as:
\begin{equation}
f_\tau(\lambda)
=
f_\tau(q)
+\frac{\tau(\lambda-q)}{(q+\tau)^2}
-\frac{\tau(\lambda-q)^2}
{(q+\tau)^2(\lambda+\tau)}.
\label{eq:csc_function_identity}
\end{equation}
We now apply Eq.~\eqref{eq:csc_function_identity}
to each eigenvalue $\lambda_j$ and sum over
$j \in \{1,\ldots,r\}$:
\begin{align}
\sum_{j=1}^{r}f_\tau(\lambda_j)
&=
\sum_{j=1}^{r}f_\tau(q)
+\frac{\tau}{(q+\tau)^2}
\sum_{j=1}^{r}(\lambda_j-q)
\nonumber\\
&\quad
-\frac{\tau}{(q+\tau)^2}
\sum_{j=1}^{r}
\frac{(\lambda_j-q)^2}{\lambda_j+\tau}.
\label{eq:csc_function_sum}
\end{align}
Because $f_\tau(q)$ is independent of $j$, we have:
\begin{equation*}
\sum_{j=1}^{r}f_\tau(q)=rf_\tau(q).
\end{equation*}
Moreover, by Eq.~\eqref{eq:csc_zero_deviation}:
\begin{equation*}
\frac{\tau}{(q+\tau)^2}
\sum_{j=1}^{r}(\lambda_j-q)=0.
\end{equation*}
Hence, Eq. (\ref{eq:csc_function_sum}) becomes:
\begin{equation}
\sum_{j=1}^{r}f_\tau(\lambda_j)
=
rf_\tau(q)
-\frac{\tau}{(q+\tau)^2}
\sum_{j=1}^{r}
\frac{(\lambda_j-q)^2}{\lambda_j+\tau}.
\label{eq:csc_spectral_decomposition}
\end{equation}

Substituting Eq.~\eqref{eq:csc_spectral_decomposition}
into the original loss in Eq.~\eqref{eq:csc_loss_proof},
we obtain:
\begin{align}
\ell_{\mathrm{csc}}
&=
1-\frac{1+\kappa}{r}
\left[
rf_\tau(q)
-\frac{\tau}{(q+\tau)^2}
\sum_{j=1}^{r}
\frac{(\lambda_j-q)^2}{\lambda_j+\tau}
\right]
\nonumber\\
&=
1-(1+\kappa)f_\tau(q)
+
\frac{(1+\kappa)\tau}{r(q+\tau)^2}
\sum_{j=1}^{r}
\frac{(\lambda_j-q)^2}{\lambda_j+\tau}.
\label{eq:csc_loss_intermediate}
\end{align}
To simplify the first term, recall that
$q=t/r$ and $\tau=\kappa/r$. Therefore:
\begin{align*}
f_\tau(q)
&=
\frac{q}{q+\tau}
=
\frac{t/r}{t/r+\kappa/r}
=
\frac{t}{t+\kappa}.
\end{align*}
Consequently:
\begin{align}
1-(1+\kappa)f_\tau(q)
&=
1-\frac{(1+\kappa)t}{t+\kappa}
\nonumber\\
&=
\frac{t+\kappa-(1+\kappa)t}{t+\kappa}
\nonumber\\
&=
\frac{t+\kappa-t-\kappa t}{t+\kappa}
\nonumber\\
&=
\frac{\kappa(1-t)}{t+\kappa}.
\label{eq:csc_alignment_term}
\end{align}
Substituting Eq.~\eqref{eq:csc_alignment_term}
into Eq.~\eqref{eq:csc_loss_intermediate} gives
the desired exact decomposition:
\begin{equation}
\ell_{\mathrm{csc}}
=
\underbrace{
\frac{\kappa(1-t)}{t+\kappa}
}_{\ell_{\mathrm{align}}}
+
\underbrace{
\frac{(1+\kappa)\tau}{r(q+\tau)^2}
\sum_{j=1}^{r}
\frac{(\lambda_j-q)^2}{\lambda_j+\tau}
}_{\ell_{\mathrm{spec}}}.
\label{eq:csc_exact_decomposition}
\end{equation}

Finally, we establish the nonnegativity of both terms.
Since $\kappa>0$ and $0\le t\le1$, we have:
\begin{equation*}
\ell_{\mathrm{align}}
=
\frac{\kappa(1-t)}{t+\kappa}
\ge0.
\end{equation*}
Similarly, since $\tau>0$, $\lambda_j\ge0$, and
$(\lambda_j-q)^2\ge0$ for every $j$, we obtain:
\begin{equation*}
\ell_{\mathrm{spec}}
=
\frac{(1+\kappa)\tau}{r(q+\tau)^2}
\sum_{j=1}^{r}
\frac{(\lambda_j-q)^2}{\lambda_j+\tau}
\ge0.
\end{equation*}
Furthermore, every coefficient and denominator in the
spectral term is strictly positive. Therefore:
\begin{equation*}
\ell_{\mathrm{spec}}=0
\iff
\lambda_j=q,\qquad \forall j=1,\ldots,r.
\end{equation*}

\subsection{Proof of Proposition \ref{prop:csc_align_bound}}\label{section_proof_csc_align_bound}

According to Eq. (\ref{eq:csc_score_upper}), we have:
\begin{equation*}
1-\ell_{\mathrm{csc}}
\le
\frac{(1+\kappa)t}{t+\kappa}
\le1.
\end{equation*}
By the assumption of the proposition:
\begin{equation*}
\ell_{\mathrm{csc}}\le\eta<1,
\end{equation*}
where we may take $0\le\eta<1$ because the CSC loss
is nonnegative. Subtracting the inequality
$\ell_{\mathrm{csc}}\le\eta$ from one gives:
\begin{equation*}
1-\ell_{\mathrm{csc}}\ge1-\eta.
\end{equation*}
Combining this inequality with
Eq.~(\ref{eq:csc_score_upper}), we obtain:
\begin{equation*}
1-\eta
\le
1-\ell_{\mathrm{csc}}
\le
\frac{(1+\kappa)t}{t+\kappa} \leq 1.
\end{equation*}
Hence:
\begin{equation}
1-\eta
\le
\frac{(1+\kappa)t}{t+\kappa}.
\label{eq:csc_alignment_proof_trace_ineq}
\end{equation}
Since $t+\kappa>0$, multiplying both sides of
Eq.~\eqref{eq:csc_alignment_proof_trace_ineq}
by $t+\kappa$ preserves the inequality:
\begin{equation*}
(1-\eta)(t+\kappa)
\le
(1+\kappa)t.
\end{equation*}
Expanding both sides yields:
\begin{align*}
&t-\eta t+\kappa-\kappa\eta
\le
t+\kappa t \nonumber \\
&\implies \kappa-\kappa\eta
\le
\kappa t+\eta t
\nonumber\\
&\implies \kappa(1-\eta)
\le
(\kappa+\eta)t.
\end{align*}
Since $\kappa+\eta>0$, division gives:
\begin{equation}
t\ge
\frac{\kappa(1-\eta)}{\kappa+\eta}.
\label{eq:csc_alignment_proof_trace_lower_bound}
\end{equation}
Thus, a sufficiently small CSC loss necessarily implies
a sufficiently large centroid energy $t$.

According to Eqs. \eqref{eq:csc_trace_floor} and (\ref{eq:csc_alignment_proof_trace_lower_bound}), we have:
\begin{align*}
\frac{1}{b}\|\b{U}-\b{V}\|_F^2
\le4(1-t) \leq \frac{4\eta(1+\kappa)}{\kappa+\eta}.
\end{align*}
This establishes the desired alignment certificate for
unit-normalized augmented views.

Finally, we show that the same bound remains valid when
the normalization floor is active. In that case, the
normalized representations satisfy:
\begin{equation*}
\|\b{u}_i\|_2\le1,
\qquad
\|\b{v}_i\|_2\le1.
\end{equation*}
Therefore:
\begin{equation*}
\|\b{U}\|_F^2\le b,
\qquad
\|\b{V}\|_F^2\le b.
\end{equation*}
Applying these inequalities to
Eq.~\eqref{eq:csc_trace_general}
gives:
\begin{align*}
t
&=
\frac{\|\b{U}\|_F^2+\|\b{V}\|_F^2}{2b}
-
\frac{\|\b{U}-\b{V}\|_F^2}{4b}
\nonumber\\
&\le
1-\frac{\|\b{U}-\b{V}\|_F^2}{4b}.
\end{align*}
Consequently:
\begin{equation*}
\frac{1}{b}\|\b{U}-\b{V}\|_F^2
\le4(1-t).
\end{equation*}
Combining this with
Eq.~\eqref{eq:csc_alignment_proof_trace_lower_bound}
yields:
\begin{align*}
\frac{1}{b}\|\b{U}-\b{V}\|_F^2
&\le4(1-t)
\le
\frac{4\eta(1+\kappa)}{\kappa+\eta}.
\end{align*}
Thus, in both the unit-normalized case and the
normalization-floor case:
\begin{equation*}
\frac{1}{b}\|\b{U}-\b{V}\|_F^2
\le
\frac{4\eta(1+\kappa)}{\kappa+\eta}.
\end{equation*}
This completes the proof.

\subsection{Proof of
Proposition~\ref{prop:csc_frame_potential}}
\label{app:csc_frame_potential_proof}

Recall that:
\begin{equation*}
\b{G}
=
\frac{1}{b}\b{C}\b{C}^\top,
\quad
\b{C}=[\b{c}_1,\ldots,\b{c}_b].
\end{equation*}
Since $\b{G}$ is symmetric and positive
semidefinite, it admits an orthogonal
eigendecomposition:
\begin{equation*}
\b{G}
=
\b{Q}\b{\Lambda}\b{Q}^\top,
\end{equation*}
where $\b{Q}^\top\b{Q}=\b{I}_p$ and
$\b{\Lambda}$ contains the nonnegative
eigenvalues of $\b{G}$.
Squaring this matrix:
\begin{align*}
\b{G}^2
&=
(\b{Q}\b{\Lambda}\b{Q}^\top)
(\b{Q}\b{\Lambda}\b{Q}^\top)
\nonumber\\
&=
\b{Q}\b{\Lambda}
(\b{Q}^\top\b{Q})
\b{\Lambda}\b{Q}^\top
=
\b{Q}\b{\Lambda}^2\b{Q}^\top.
\end{align*}
Taking the trace and applying its
cyclic property gives:
\begin{align}
\operatorname{tr}(\b{G}^2)
&=
\operatorname{tr}
(\b{Q}\b{\Lambda}^2\b{Q}^\top)
=
\operatorname{tr}
(\b{\Lambda}^2\b{Q}^\top\b{Q})
\nonumber\\
&=
\operatorname{tr}(\b{\Lambda}^2)
=
\sum_{j=1}^{r}\lambda_j^2.
\label{eq:csc_frame_spectral_proof}
\end{align}
The final equality follows because
$\operatorname{rank}(\b{G})\le r=\min(b,p)$,
so the remaining eigenvalues are zero.

We now derive an alternative expression
using the centroid vectors.
Substituting the definition of $\b{G}$ gives:
\begin{align}
\operatorname{tr}(\b{G}^2)
&=
\operatorname{tr}
\left[
\left(
\frac{1}{b}\b{C}\b{C}^\top
\right)^2
\right]
\nonumber\\
&=
\frac{1}{b^2}
\operatorname{tr}
(\b{C}\b{C}^\top\b{C}\b{C}^\top)
\nonumber\\
&\overset{(a)}{=}
\frac{1}{b^2}
\operatorname{tr}
(\b{C}^\top\b{C}\b{C}^\top\b{C})
\nonumber\\
&=
\frac{1}{b^2}
\operatorname{tr}
\left[
(\b{C}^\top\b{C})^2
\right].
\label{eq:csc_frame_gram_proof}
\end{align}
where $(a)$ follows from
the cyclic property of the trace.

Let:
\begin{equation*}
\b{K}=\b{C}^\top\b{C}
\in\mathbb{R}^{b\times b}.
\end{equation*}
This is the Gram matrix of the centroids,
whose $(i,j)$-th entry is:
\begin{equation*}
K_{ij}=\b{c}_i^\top\b{c}_j.
\end{equation*}
Since $\b{K}$ is symmetric:
\begin{align}
\operatorname{tr}(\b{K}^2)
&=
\sum_{i=1}^{b}(\b{K}^2)_{ii}
=
\sum_{i=1}^{b}\sum_{j=1}^{b}
K_{ij}K_{ji}
\nonumber\\
&=
\sum_{i=1}^{b}\sum_{j=1}^{b}
K_{ij}^2
=
\sum_{i=1}^{b}\sum_{j=1}^{b}
(\b{c}_i^\top\b{c}_j)^2.
\end{align}
Combining this expression with
Eqs.~\eqref{eq:csc_frame_spectral_proof}
and~\eqref{eq:csc_frame_gram_proof}, we obtain:
\begin{equation*}
\operatorname{tr}(\b{G}^2)
=
\sum_{j=1}^{r}\lambda_j^2
=
\frac{1}{b^2}
\sum_{i=1}^{b}\sum_{j=1}^{b}
(\b{c}_i^\top\b{c}_j)^2.
\end{equation*}

We next establish the lower bound.
Consider the vectors:
\begin{equation*}
\b{a}
=
[\lambda_1,\ldots,\lambda_r]^\top,
\quad
\b{e}
=
[1,\ldots,1]^\top
\in\mathbb{R}^{r}.
\end{equation*}
By the Cauchy--Schwarz inequality, we have:
\begin{equation*}
(\b{a}^\top\b{e})^2
\le
\|\b{a}\|_2^2\, \|\b{e}\|_2^2.
\end{equation*}
Substituting their components:
\begin{equation*}
\left(
\sum_{j=1}^{r}\lambda_j
\right)^2
\le
\left(
\sum_{j=1}^{r}\lambda_j^2
\right)
\left(
\sum_{j=1}^{r}1^2
\right).
\end{equation*}
Because:
\begin{equation*}
\sum_{j=1}^{r}\lambda_j=t,
\qquad
\sum_{j=1}^{r}1^2=r,
\end{equation*}
we obtain:
\begin{equation*}
t^2\le r\sum_{j=1}^{r}\lambda_j^2.
\end{equation*}
Dividing by $r$ gives:
\begin{equation*}
\operatorname{tr}(\b{G}^2)
=
\sum_{j=1}^{r}\lambda_j^2
\ge\frac{t^2}{r}.
\end{equation*}

Equality in Cauchy--Schwarz holds
if and only if $\b{a}$ and $\b{e}$
are linearly dependent. Thus:
\begin{equation*}
\b{a}=\alpha\b{e},
\end{equation*}
for some scalar $\alpha$, implying:
\begin{equation*}
\lambda_1=\cdots=\lambda_r=\alpha.
\end{equation*}
Since the eigenvalues sum to $t$:
\begin{equation*}
r\alpha=t
\quad\Longrightarrow\quad
\alpha=\frac{t}{r}.
\end{equation*}
Therefore, equality holds exactly when:
\begin{equation*}
\lambda_1=\cdots=\lambda_r=\frac{t}{r}.
\end{equation*}
Finally, we establish the connection
to the CSC objective at fixed energy.

Recall that:
\begin{equation*}
f_\tau(\lambda)
=
\frac{\lambda}{\lambda+\tau},
\qquad \tau>0.
\end{equation*}
Its second derivative is:
\begin{equation*}
f_\tau''(\lambda)
=
-\frac{2\tau}{(\lambda+\tau)^3}<0.
\end{equation*}
Hence, $f_\tau$ is strictly concave.
Applying Jensen's inequality:
\begin{align}
\frac{1}{r}
\sum_{j=1}^{r}f_\tau(\lambda_j)
&\le
f_\tau
\left(
\frac{1}{r}
\sum_{j=1}^{r}\lambda_j
\right)
=
f_\tau\left(\frac{t}{r}\right).
\end{align}
Equality holds if and only if:
\begin{equation}
\lambda_1=\cdots=\lambda_r=\frac{t}{r}.
\end{equation}
Therefore, for fixed $t$, the CSC
spectral score is maximized by the same
flat eigenvalue distribution that
minimizes $\operatorname{tr}(\b{G}^2)$.
Since minimizing the CSC loss is
equivalent to maximizing its spectral
score, the final claim follows.

We can further express the frame potential in terms
of squared deviations from the mean eigenvalue.
Recall that:
\begin{equation*}
q=\frac{t}{r},
\qquad
t=\sum_{j=1}^{r}\lambda_j.
\end{equation*}
Expanding the squared deviations gives:
\begin{align*}
\sum_{j=1}^{r}(\lambda_j-q)^2
&=
\sum_{j=1}^{r}
(\lambda_j^2-2q\lambda_j+q^2)
\nonumber\\
&=
\sum_{j=1}^{r}\lambda_j^2
-2q\sum_{j=1}^{r}\lambda_j
+rq^2
\nonumber\\
&=
\sum_{j=1}^{r}\lambda_j^2
-2qt+rq^2
\nonumber\\
&=
\sum_{j=1}^{r}\lambda_j^2
-\frac{2t^2}{r}
+\frac{t^2}{r}
\nonumber\\
&=
\sum_{j=1}^{r}\lambda_j^2
-\frac{t^2}{r}.
\end{align*}
Rearranging, and using the spectral identity
$\operatorname{tr}(\b{G}^2)
=\sum_{j=1}^{r}\lambda_j^2$, yields:
\begin{equation*}
\operatorname{tr}(\b{G}^2)
=
\frac{t^2}{r}
+
\sum_{j=1}^{r}(\lambda_j-q)^2.
\end{equation*}

\subsection{Proof of
Proposition~\ref{prop:csc_zero_loss_geometry}}
\label{app:csc_zero_loss_geometry_proof}

Recall the CSC loss:
\begin{equation*}
\ell_{\mathrm{csc}}
=
1-\frac{1+\kappa}{r}
\sum_{j=1}^{r}
\frac{\lambda_j}{\lambda_j+\kappa/r}.
\end{equation*}
By Jensen's inequality, as described in Eq. (\ref{eq:csc_alignment_proof_score_bound}), we have:
\begin{equation*}
1-\ell_{\mathrm{csc}}
\overset{(\ref{eq:csc_alignment_proof_score_bound})}{\le}
\frac{(1+\kappa)t}{t+\kappa}
\overset{(a)}{\le} 1,
\end{equation*}
where $(a)$ follows from $t\le1$.
If $\ell_{\mathrm{csc}}=0$, then:
\begin{equation*}
1
\le
\frac{(1+\kappa)t}{t+\kappa}
\le1.
\end{equation*}
Therefore, equality must hold, giving $t=1$.

Recall that:
\begin{equation*}
t=
\frac{1}{b}
\sum_{i=1}^{b}\|\b{c}_i\|_2^2,
\qquad
\|\b{c}_i\|_2\le1.
\end{equation*}
Since the average of these squared
norms equals one and each is at most
one, every centroid must satisfy:
\begin{equation*}
\|\b{c}_i\|_2=1.
\end{equation*}
Moreover:
\begin{equation*}
\b{c}_i=
\frac{\b{u}_i+\b{v}_i}{2}.
\end{equation*}
Using the parallelogram identity, we have:
\begin{equation*}
\|\b{c}_i\|_2^2
=
\frac{
\|\b{u}_i\|_2^2+
\|\b{v}_i\|_2^2
}{2}
-
\frac14
\|\b{u}_i-\b{v}_i\|_2^2.
\end{equation*}
Because both augmented views have
norm at most one, equality
$\|\b{c}_i\|_2=1$ requires:
\begin{equation*}
\b{u}_i=\b{v}_i=\b{c}_i,
\qquad
\|\b{u}_i\|_2=
\|\b{v}_i\|_2=1.
\end{equation*}
Furthermore, equality in Jensen's
inequality requires all $r$ eigenvalues
to be identical:
\begin{equation*}
\lambda_1=\cdots=\lambda_r=\frac{t}{r}
=\frac{1}{r}.
\end{equation*}
Consequently:
\begin{align*}
\operatorname{tr}(\b{G}^2)
&=
\sum_{j=1}^{r}\lambda_j^2
=
r\left(\frac{1}{r}\right)^2
=
\frac{1}{r}.
\end{align*}

Recall that the frame-potential identity is:
\begin{equation*}
\operatorname{tr}(\b{G}^2)
=
\frac{1}{b^2}
\sum_{i=1}^{b}\sum_{j=1}^{b}
(\b{c}_i^\top\b{c}_j)^2.
\end{equation*}
Comparing these two equations gives:
\begin{equation*}
\frac{1}{r}
=
\frac{1}{b^2}
\sum_{i=1}^{b}\sum_{j=1}^{b}
(\b{c}_i^\top\b{c}_j)^2
\implies 
\sum_{i=1}^{b}\sum_{j=1}^{b}
(\b{c}_i^\top\b{c}_j)^2
=
\frac{b^2}{r}.
\end{equation*}
Separating diagonal and off-diagonal
terms:
\begin{align}\label{equation_b2_r_sum_ct_c2_sum_ct_c2}
\frac{b^2}{r}
&=
\sum_{i=1}^{b}
(\b{c}_i^\top\b{c}_i)^2
+
\sum_{i\ne j}
(\b{c}_i^\top\b{c}_j)^2.
\end{align}

Since $\|\b{c}_i\|_2=1$:
\begin{equation*}
(\b{c}_i^\top\b{c}_i)^2
=
\|\b{c}_i\|_2^4=1.
\end{equation*}
Therefore:
\begin{equation*}
\sum_{i=1}^{b}
(\b{c}_i^\top\b{c}_i)^2=b.
\end{equation*}
Thus, Eq. (\ref{equation_b2_r_sum_ct_c2_sum_ct_c2}) becomes:
\begin{equation}
\sum_{i\ne j}
(\b{c}_i^\top\b{c}_j)^2
=
\frac{b^2}{r}-b.
\end{equation}

We now consider the two possible
relationships between batch size
and embedding dimension.

\textit{Case 1: $b\le p$.}
In this case:
\begin{equation}
r=\min(b,p)=b.
\end{equation}
Hence:
\begin{align}
\sum_{i\ne j}
(\b{c}_i^\top\b{c}_j)^2
&=
\frac{b^2}{b}-b
=0.
\end{align}
Since each squared inner product is
nonnegative, their sum can equal
zero only when every term is zero.
Thus:
\begin{equation}
\b{c}_i^\top\b{c}_j=0,
\qquad i\ne j.
\end{equation}
Therefore, all distinct centroids
are mutually orthogonal.

\textit{Case 2: $b>p$.}
Here:
\begin{equation}
r=\min(b,p)=p.
\end{equation}
Since all $p$ eigenvalues of $\b{G}$
equal $1/p$, its eigendecomposition is:
\begin{align*}
\b{G}
&=
\b{Q}
\left(\frac{1}{p}\b{I}_p\right)
\b{Q}^\top
=
\frac{1}{p}\b{Q}\b{Q}^\top
=
\frac{1}{p}\b{I}_p.
\end{align*}
By the definition of $\b{G}$ in Eq. (\ref{eq:csc_second_moment}), this becomes:
\begin{equation*}
\b{C}\b{C}^\top
=
\frac{b}{p}\b{I}_p.
\end{equation*}

Together with:
\begin{equation*}
\|\b{c}_i\|_2=1,
\qquad i=1,\ldots,b,
\end{equation*}
this is precisely the defining
condition of a unit-norm tight frame.

Finally, the off-diagonal squared
inner products satisfy:
\begin{equation*}
\sum_{i\ne j}
(\b{c}_i^\top\b{c}_j)^2
=
\frac{b^2}{p}-b.
\end{equation*}
Since $b>p$, this quantity is
strictly positive. Therefore,
mutual orthogonality of all
centroids is impossible.

\subsection{Proof of Lemma \ref{lemma_pairwise_distance_identity}}\label{section_proof_lemma_pairwise_distance_identity}

Recall that the centered covariance matrix is:
\begin{equation*}
\b{\Sigma}
=
\frac{1}{b}\sum_{i=1}^{b}
(\b{c}_i-\b{\mu})(\b{c}_i-\b{\mu})^\top,
\end{equation*}
where:
\begin{equation*}
\b{\mu}=\frac{1}{b}\sum_{i=1}^{b}\b{c}_i.
\end{equation*}
Taking the trace and expanding gives:
\begin{align}
\operatorname{tr}(\b{\Sigma})
&=
\frac{1}{b}\sum_{i=1}^{b}
\|\b{c}_i-\b{\mu}\|_2^2
\nonumber\\
&=
\frac{1}{b}\sum_{i=1}^{b}
\left(
\|\b{c}_i\|_2^2
-2\b{c}_i^\top\b{\mu}
+\|\b{\mu}\|_2^2
\right)
\nonumber\\
&\overset{(a)}{=}
\frac{1}{b}\sum_{i=1}^{b}\|\b{c}_i\|_2^2
-\|\b{\mu}\|_2^2,
\label{eq:csc_centered_trace_proof}
\end{align}
where $(a)$ is because $(1/b)\sum_{i=1}^{b}\b{c}_i = \b{\mu}$.

On the other hand, expanding the sum of
pairwise squared distances yields:
\begin{align*}
\sum_{i,j=1}^{b}
\|&\b{c}_i-\b{c}_j\|_2^2
\nonumber\\
&=
\sum_{i,j=1}^{b}
\left(
\|\b{c}_i\|_2^2
+\|\b{c}_j\|_2^2
-2\b{c}_i^\top\b{c}_j
\right)
\nonumber\\
&=
2b\sum_{i=1}^{b}\|\b{c}_i\|_2^2
-2\sum_{i,j=1}^{b}\b{c}_i^\top\b{c}_j
\nonumber\\
&=
2b\sum_{i=1}^{b}\|\b{c}_i\|_2^2
-2\left\|\sum_{i=1}^{b}\b{c}_i\right\|_2^2
\nonumber\\
&\overset{(a)}{=}
2b\sum_{i=1}^{b}\|\b{c}_i\|_2^2
-2b^2\|\b{\mu}\|_2^2,
\end{align*}
where $(a)$ is because $(1/b)\sum_{i=1}^{b}\b{c}_i = \b{\mu}$.
Dividing both sides by $2b^2$:
\begin{align*}
\frac{1}{2b^2}\sum_{i,j=1}^{b}
\|\b{c}_i-\b{c}_j\|_2^2
&=
\frac{1}{b}\sum_{i=1}^{b}
\|\b{c}_i\|_2^2
-\|\b{\mu}\|_2^2
\nonumber\\
&\overset{
(\ref{eq:csc_centered_trace_proof})
}{=}
\operatorname{tr}(\b{\Sigma}).
\end{align*}
Therefore:
\begin{equation*}
\operatorname{tr}(\b{\Sigma})
=
\frac{1}{2b^2}\sum_{i,j=1}^{b}
\|\b{c}_i-\b{c}_j\|_2^2.
\end{equation*}

\subsection{Proof of Proposition~\ref{prop:csc_finite_rank}}
\label{app:csc_finite_rank_proof}
Recall the definition of the CSC objective:
\begin{equation*}
\ell_{\mathrm{csc}}
=
1-\frac{1+\kappa}{r}
\sum_{j=1}^{r}
\frac{\lambda_j}{\lambda_j+\kappa/r},
\end{equation*}
where $\kappa>0$, $r=\min(b,p)$, and
$\lambda_1,\ldots,\lambda_r$ are the potentially
nonzero eigenvalues of $\b{G}$.
Let:
\begin{equation*}
k:=\operatorname{rank}(\b{G}).
\end{equation*}
Since $\b{G}$ is positive semidefinite, its
eigenvalues are nonnegative, and exactly $k$
of them are strictly positive.
We first consider $k\ge1$. Without loss of
generality, let $\lambda_1,\ldots,\lambda_k>0$,
while the remaining $r-k$ eigenvalues are zero.
Recall that:
\begin{equation*}
f_\tau(\lambda)
=
\frac{\lambda}{\lambda+\tau},
\qquad
\tau=\frac{\kappa}{r}>0.
\end{equation*}
Since $f_\tau(0)=0$, the CSC score becomes:
\begin{equation*}
1-\ell_{\mathrm{csc}}
=
\frac{1+\kappa}{r}
\sum_{j=1}^{k}f_\tau(\lambda_j).
\end{equation*}
The function $f_\tau$ is strictly concave on
$[0,\infty)$ because:
\begin{equation*}
f_\tau''(\lambda)
=
-\frac{2\tau}{(\lambda+\tau)^3}<0.
\end{equation*}
Therefore, applying Jensen's inequality to
the $k$ positive eigenvalues gives:
\begin{equation*}
\frac{1}{k}\sum_{j=1}^{k}f_\tau(\lambda_j)
\le
f_\tau\left(
\frac{1}{k}\sum_{j=1}^{k}\lambda_j
\right).
\end{equation*}
Since the remaining eigenvalues are zero, their
sum equals the total centroid energy:
\begin{equation*}
\sum_{j=1}^{k}\lambda_j
=
\sum_{j=1}^{r}\lambda_j
=
t.
\end{equation*}
Consequently:
\begin{align*}
1-\ell_{\mathrm{csc}}
&=
\frac{1+\kappa}{r}
\sum_{j=1}^{k}f_\tau(\lambda_j)\\
&\le
\frac{(1+\kappa)k}{r}
f_\tau\left(\frac{t}{k}\right)\\
&=
\frac{(1+\kappa)k}{r}
\frac{t/k}{t/k+\tau}\\
&=
\frac{(1+\kappa)kt}{r(t+k\tau)}.
\end{align*}
Recall that:
\begin{equation*}
t=\frac{1}{b}\sum_{i=1}^{b}
\|\b{c}_i\|_2^2\le1,
\end{equation*}
because $\|\b{c}_i\|_2\le1$ for every
instance.
For any fixed $a>0$, the function
$t/(t+a)$ is strictly increasing for
$t\ge0$, since:
\begin{equation*}
\frac{d}{dt}\left(\frac{t}{t+a}\right)
=
\frac{a}{(t+a)^2}>0.
\end{equation*}
Using $t\le1$ and $k\tau>0$, we obtain:
\begin{equation*}
\frac{t}{t+k\tau}
\le
\frac{1}{1+k\tau}.
\end{equation*}
Therefore:
\begin{align*}
1-\ell_{\mathrm{csc}}
&\le
\frac{(1+\kappa)k}{r(1+k\tau)}\\
&=
\frac{(1+\kappa)k}{r+\kappa k},
\end{align*}
where the last equality uses
$\tau=\kappa/r$.
Subtracting both sides from one yields:
\begin{align*}
\ell_{\mathrm{csc}}
&\ge
1-\frac{(1+\kappa)k}{r+\kappa k}\\
&=
\frac{r+\kappa k-(1+\kappa)k}
{r+\kappa k}\\
&=
\frac{r-k}{r+\kappa k}.
\end{align*}
If $k=0$, then $\b{G}=\b{0}$ and all
eigenvalues vanish. Hence,
$\ell_{\mathrm{csc}}=1$, which equals
$(r-k)/(r+\kappa k)=1$.
Thus, the bound holds for every
$0\le k\le r$:
\begin{equation*}
\ell_{\mathrm{csc}}
\ge
\frac{r-k}{r+\kappa k}.
\end{equation*}
We next establish the finite-loss rank
certificate. Suppose:
\begin{equation*}
0\le\ell_{\mathrm{csc}}\le\eta<1.
\end{equation*}
Combining this assumption with the
previous bound gives:
\begin{equation*}
\frac{r-k}{r+\kappa k}
\le
\ell_{\mathrm{csc}}
\le\eta.
\end{equation*}
Since $r+\kappa k>0$, multiplying by
this denominator preserves the inequality:
\begin{equation*}
r-k\le\eta(r+\kappa k).
\end{equation*}
Expanding and collecting the terms
involving $k$ gives:
\begin{align*}
r-k
&\le\eta r+\eta\kappa k \implies
r(1-\eta) \le k(1+\kappa\eta).
\end{align*}
Because $1+\kappa\eta>0$, division gives:
\begin{equation*}
\operatorname{rank}(\b{G})=k
\ge
\frac{r(1-\eta)}{1+\kappa\eta}.
\end{equation*}
We now relate the rank of $\b{G}$ to
the rank of the centered covariance
$\b{\Sigma}$.
Recall that:
\begin{equation*}
\b{\Sigma}
=
\b{G}-\b{\mu}\b{\mu}^\top.
\end{equation*}
Equivalently:
\begin{equation*}
\b{G}
=
\b{\Sigma}+\b{\mu}\b{\mu}^\top.
\end{equation*}
By the subadditivity of matrix rank~\cite{horn2012matrix}, for any two matrices of the same size,
the rank of their sum does not exceed
the sum of their ranks. Therefore:
\begin{align*}
\operatorname{rank}(\b{G})
&\le
\operatorname{rank}(\b{\Sigma})
+
\operatorname{rank}(\b{\mu}\b{\mu}^\top)\\
&\le
\operatorname{rank}(\b{\Sigma})+1,
\end{align*}
where the second inequality follows
because an outer product has rank at
most one.
Rearranging:
\begin{equation*}
\operatorname{rank}(\b{\Sigma})
\ge
\operatorname{rank}(\b{G})-1.
\end{equation*}
Finally, we prove the noncollapse
condition. Suppose $r>1$ and:
\begin{equation*}
\ell_{\mathrm{csc}}
<
\frac{r-1}{r+\kappa}.
\end{equation*}
If $\b{\Sigma}=\b{0}$, then:
\begin{equation*}
\b{G}=\b{\mu}\b{\mu}^\top,
\end{equation*}
which implies $\operatorname{rank}(\b{G})
\le1$.
If this rank is one, the loss-gap bound
gives:
\begin{equation*}
\ell_{\mathrm{csc}}
\ge
\frac{r-1}{r+\kappa}.
\end{equation*}
If the rank is zero, the CSC loss equals
one, which also satisfies this lower bound.
Both cases contradict the assumed strict
inequality.
Therefore, $\b{\Sigma}$ cannot be zero:
\begin{equation*}
\ell_{\mathrm{csc}}
<
\frac{r-1}{r+\kappa}
\quad\Longrightarrow\quad
\b{\Sigma}\ne\b{0}.
\end{equation*}
This completes the proof.

\subsection{Proof of
Proposition~\ref{prop:csc_centered_nocollapse}}
\label{app:csc_centered_nocollapse_proof}
Suppose $r=\min(b,p)>1$ and
$\ell_{\mathrm{csc}}=0$.
According to Proposition~\ref{prop:csc_optimum},
zero CSC loss is attained only when the
centroids have unit norm and the $r$
potentially nonzero eigenvalues of
$\b{G}$ are all equal to $1/r$.
Therefore:
\begin{equation*}
\lambda_1=\cdots=\lambda_r=\frac{1}{r},
\end{equation*}
and consequently:
\begin{equation*}
\operatorname{rank}(\b{G})=r,
\qquad
\operatorname{tr}(\b{G})=1,
\qquad
\lambda_{\max}(\b{G})=\frac{1}{r}.
\end{equation*}
We first establish the rank lower bound.
Recall that the centered covariance satisfies:
\begin{equation*}
\b{\Sigma}
=
\b{G}-\b{\mu}\b{\mu}^\top,
\end{equation*}
or equivalently:
\begin{equation*}
\b{G}
=
\b{\Sigma}+\b{\mu}\b{\mu}^\top.
\end{equation*}
Using the subadditivity of matrix rank:
\begin{align*}
\operatorname{rank}(\b{G})
&\le
\operatorname{rank}(\b{\Sigma})
+
\operatorname{rank}(\b{\mu}\b{\mu}^\top)\\
&\le
\operatorname{rank}(\b{\Sigma})+1.
\end{align*}
Since $\operatorname{rank}(\b{G})=r$,
we obtain:
\begin{equation}
\operatorname{rank}(\b{\Sigma})
\ge r-1.
\end{equation}
Thus, at zero CSC loss, centering the
representations can remove at most one
dimension from their rank.
We next derive the lower bound on the
total centered variance.
By the definition of $\b{\Sigma}$:
\begin{equation*}
\b{\Sigma}
=
\b{G}-\b{\mu}\b{\mu}^\top
\succeq\b{0}.
\end{equation*}
Hence:
\begin{equation*}
\b{\mu}\b{\mu}^\top
\preceq\b{G},
\end{equation*}
where $\preceq$ denotes the
positive-semidefinite ordering.
This means that for every
$\b{x}\in\mathbb{R}^p$:
\begin{equation*}
\b{x}^\top
\b{\mu}\b{\mu}^\top\b{x}
\le
\b{x}^\top\b{G}\b{x}, \quad \forall \b{x}\in\mathbb{R}^p.
\end{equation*}
We use this inequality to bound
$\|\b{\mu}\|_2^2$.
If $\b{\mu}=\b{0}$, then the bound
$\|\b{\mu}\|_2^2\le1/r$ holds trivially.
Otherwise, define the unit vector:
\begin{equation*}
\b{x}:=
\frac{\b{\mu}}{\|\b{\mu}\|_2}.
\end{equation*}
Substitution gives:
\begin{equation*}
\b{x}^\top
\b{\mu}\b{\mu}^\top\b{x}
=
(\b{x}^\top\b{\mu})^2
=
\|\b{\mu}\|_2^2.
\end{equation*}
Moreover, because $\b{G}$ is symmetric
and positive semidefinite, the Rayleigh--Ritz theorem~\cite{horn2012matrix} yields:
\begin{equation*}
\b{x}^\top\b{G}\b{x}
\le
\lambda_{\max}(\b{G})
\|\b{x}\|_2^2.
\end{equation*}
Since $\|\b{x}\|_2=1$ and
$\lambda_{\max}(\b{G})=1/r$, we obtain:
\begin{align*}
\|\b{\mu}\|_2^2
&=
\b{x}^\top\b{\mu}\b{\mu}^\top\b{x}\le
\b{x}^\top\b{G}\b{x}\le
\lambda_{\max}(\b{G})=
\frac{1}{r}.
\end{align*}
Therefore, in all cases:
\begin{equation}
\|\b{\mu}\|_2^2\le\frac{1}{r}.
\end{equation}
Taking the trace of the centered
covariance gives:
\begin{align*}
\operatorname{tr}(\b{\Sigma})
&=
\operatorname{tr}
(\b{G}-\b{\mu}\b{\mu}^\top)\\
&=
\operatorname{tr}(\b{G})
-\operatorname{tr}(\b{\mu}\b{\mu}^\top)\\
&=
\operatorname{tr}(\b{G})
-\|\b{\mu}\|_2^2.
\end{align*}
Using $\operatorname{tr}(\b{G})=1$
and $\|\b{\mu}\|_2^2\le1/r$, we obtain the
trace lower bound:
\begin{equation}\label{equation_tr_Sigma_gew_1_1r}
\operatorname{tr}(\b{\Sigma})
\ge1-\frac{1}{r}.
\end{equation}
This establishes the second claim.
Finally, we obtain the average
pairwise-distance bound.
By the pairwise-distance identity in
Eq.~(\ref{eq:csc_pairwise_diversity}):
\begin{equation*}
\operatorname{tr}(\b{\Sigma})
=
\frac{1}{2b^2}
\sum_{i,j=1}^{b}
\|\b{c}_i-\b{c}_j\|_2^2.
\end{equation*}
Combining this identity with Eq. (\ref{equation_tr_Sigma_gew_1_1r}) provides:
\begin{equation*}
\frac{1}{2b^2}
\sum_{i,j=1}^{b}
\|\b{c}_i-\b{c}_j\|_2^2
\ge
1-\frac{1}{r}.
\end{equation*}
Multiplying both sides by two gives:
\begin{equation}
\frac{1}{b^2}
\sum_{i,j=1}^{b}
\|\b{c}_i-\b{c}_j\|_2^2
\ge
2\left(1-\frac{1}{r}\right).
\end{equation}
Since $r>1$, the right-hand side is
strictly positive:
\begin{equation*}
2\left(1-\frac{1}{r}\right)>0.
\end{equation*}
Therefore, the average squared distance
between the centroids cannot be zero.
In particular, the centroids cannot
all be identical, which excludes
complete centroid collapse at zero
CSC loss.
This completes the proof.

\subsection{Proof of
Proposition~\ref{prop:csc_finite_variance}}
\label{app:csc_finite_variance_proof}
Recall Eq. (\ref{eq:csc_centered_cov}), i.e., the decomposition of the uncentered
second-moment matrix:
\begin{equation*}
\b{G}
=
\b{\mu}\b{\mu}^\top+\b{\Sigma},
\end{equation*}
where $\b{\Sigma}$ is the centered
covariance matrix.
For convenience, define:
\begin{equation*}
\b{A}:=\b{\mu}\b{\mu}^\top,
\qquad
\b{H}:=\b{\Sigma}.
\end{equation*}
Both matrices are positive semidefinite:
\begin{equation*}
\b{A}\succeq\b{0},
\qquad
\b{H}\succeq\b{0},
\qquad
\b{G}=\b{A}+\b{H}.
\end{equation*}
We will use these properties to bound
the CSC score in terms of the total
centered variance $\operatorname{tr}(\b{H})$.

Recall that, according to Eq. (\ref{eq:csc_matrix_loss}), the CSC objective can be written in matrix form:
\begin{equation}\label{equation_1_minus_l_csc_1_plus_k_r_tr_G_G_tau_I_inv}
1-\ell_{\mathrm{csc}}
=
\frac{1+\kappa}{r}
\operatorname{tr}\left[
\b{G}(\b{G}+\tau\b{I}_p)^{-1}
\right],
\end{equation}
where $\tau=\kappa/r>0$.
Define the scalar-valued matrix function:
\begin{equation}\label{equation_hX_tr_X_X_I_inv}
h(\b{X})
:=
\operatorname{tr}\left[
\b{X}(\b{X}+\tau\b{I}_p)^{-1}
\right],
\end{equation}
for any symmetric positive-semidefinite
matrix $\b{X}\in\mathbb{R}^{p\times p}$.
Thus, Eq. (\ref{equation_1_minus_l_csc_1_plus_k_r_tr_G_G_tau_I_inv}) can be stated as:
\begin{equation}\label{equation_1_minus_l_csc_1_plus_k_r_hG}
1-\ell_{\mathrm{csc}}
=
\frac{1+\kappa}{r}
h(\b{G}).
\end{equation}

We first establish an upper bound on
$h(\b{G})$ using the decomposition
$\b{G}=\b{A}+\b{H}$.
Consider the matrix path:
\begin{equation}\label{equation_X_A_sH}
\b{X}(s):=\b{A}+s\b{H},
\qquad s\in[0,1].
\end{equation}
At the endpoints, we have:
\begin{equation*}
\b{X}(0)=\b{A},
\qquad
\b{X}(1)=\b{G}.
\end{equation*}
Define:
\begin{equation}\label{equation_B_A_sH_tauI}
\b{B}(s)
:= \b{X}(s) + \tau\b{I}_p =
\b{A}+s\b{H}+\tau\b{I}_p.
\end{equation}
Because $\b{A}\succeq\b{0}$,
$\b{H}\succeq\b{0}$, $s\ge0$, and
$\tau>0$, we have:
\begin{equation*}
\b{B}(s)\succeq\tau\b{I}_p\succ\b{0}.
\end{equation*}
Thus, $\b{B}(s)$ is positive definite
and invertible for every $s\in[0,1]$.

Using the decomposition:
\begin{equation*}
\b{X}=(\b{X}+\tau\b{I}_p)-\tau\b{I}_p,
\end{equation*}
and the invertibility of
$\b{X}+\tau\b{I}_p$ for
$\b{X}\succeq\b{0}$ and $\tau>0$,
we obtain:
\begin{align}
\b{X}(\b{X}&+\tau\b{I}_p)^{-1} \nonumber \\
&=
\left[
(\b{X}+\tau\b{I}_p)-\tau\b{I}_p
\right]
(\b{X}+\tau\b{I}_p)^{-1} \nonumber \\
&=
\b{I}_p-\tau(\b{X}+\tau\b{I}_p)^{-1}. \label{equation_X_X_tauI_inv_I_tau_X_tauI_inv}
\end{align}
Using the above formula, we can rewrite $h$ along this path:
\begin{align*}
h(\b{X})
&\overset{(\ref{equation_hX_tr_X_X_I_inv})}{=} 
\operatorname{tr}\left[
\b{X}(\b{X}+\tau\b{I}_p)^{-1}
\right]
\\
&\overset{(\ref{equation_X_X_tauI_inv_I_tau_X_tauI_inv})}{=} 
\operatorname{tr}\left[
\b{I}_p-\tau(\b{X}+\tau\b{I}_p)^{-1}
\right]\\
&\overset{(\ref{equation_B_A_sH_tauI})}{=}
\operatorname{tr}\left[
\b{I}_p-\tau\b{B}(s)^{-1}
\right]\\
&=
\operatorname{tr}[\b{I}_p]-\tau\operatorname{tr}
\left[\b{B}(s)^{-1}\right] = p-\tau\operatorname{tr}
\left[\b{B}(s)^{-1}\right].
\end{align*}
We now differentiate with respect to $s$.
For an invertible differentiable matrix
$\b{B}(s)$, the derivative of its inverse
is:
\begin{equation*}
\frac{d}{ds}\b{B}(s)^{-1}
=
-\b{B}(s)^{-1}
\frac{d\b{B}(s)}{ds}
\b{B}(s)^{-1}.
\end{equation*}
Since:
\begin{equation*}
\frac{d\b{B}(s)}{ds}
=
\b{H},
\end{equation*}
we obtain:
\begin{equation*}
\frac{d}{ds}\b{B}(s)^{-1}
=
-\b{B}(s)^{-1}
\b{H}
\b{B}(s)^{-1}.
\end{equation*}
Therefore:
\begin{align*}
\frac{d}{ds}h(\b{A}+s\b{H})
&=
-\tau\frac{d}{ds}
\operatorname{tr}\left[\b{B}(s)^{-1}\right]\\
&=
\tau\operatorname{tr}\left[
\b{B}(s)^{-1}\b{H}\b{B}(s)^{-1}
\right]\\
&\overset{(a)}{=}
\tau\operatorname{tr}\left[
\b{B}(s)^{-2}\b{H}
\right],
\end{align*}
where $(a)$ follows from the cyclic property of the trace.

We next bound this derivative.
Because $\b{B}(s)\succeq\tau\b{I}_p$,
all eigenvalues of $\b{B}(s)$ are at
least $\tau$.
Consequently, all eigenvalues of
$\b{B}(s)^{-2}$ are at most
$1/\tau^2$, giving:
\begin{equation*}
\b{0}\preceq
\b{B}(s)^{-2}
\preceq
\frac{1}{\tau^2}\b{I}_p.
\end{equation*}
Since $\b{H}\succeq\b{0}$, taking the
trace of the product with $\b{H}$
preserves the corresponding scalar
inequality:
\begin{equation*}
0\le
\operatorname{tr}\left[
\b{B}(s)^{-2}\b{H}
\right]
\le
\frac{1}{\tau^2}
\operatorname{tr}(\b{H}).
\end{equation*}
Multiplying by $\tau>0$:
\begin{equation}\label{equation_o_dhA_sH_ds_lew_1_tau_tr_H}
0\le
\frac{d}{ds}h(\b{A}+s\b{H})
\le
\frac{1}{\tau}
\operatorname{tr}(\b{H}).
\end{equation}
The right-hand side is independent
of $s$.
Integrating over $s\in[0,1]$ gives:
\begin{align*}
h(\b{A}+\b{H})-h(\b{A})
&=
\int_0^1
\frac{d}{ds}h(\b{A}+s\b{H})\,ds\\
&\overset{(\ref{equation_o_dhA_sH_ds_lew_1_tau_tr_H})}{\le}
\int_0^1
\frac{1}{\tau}\operatorname{tr}(\b{H})\,ds\\
&=
\frac{1}{\tau}\operatorname{tr}(\b{H}).
\end{align*}
Using $\b{A}+\b{H}=\b{G}$ and
$\b{H}=\b{\Sigma}$, we obtain:
\begin{equation}\label{equation_hG_leq_hA_itau_tr_Sigma}
h(\b{G})
\le
h(\b{A})
+\frac{1}{\tau}
\operatorname{tr}(\b{\Sigma}).
\end{equation}

It remains to bound $h(\b{A})$.
Recall that $\b{A}=\b{\mu}\b{\mu}^\top$
has one potentially nonzero eigenvalue,
$\|\b{\mu}\|_2^2$, with all remaining
eigenvalues equal to zero.
Using the eigendecomposition
$\b{A}=\b{Q}\b{\Lambda}\b{Q}^\top$
in Eq.~(\ref{equation_hX_tr_X_X_I_inv}),
and applying the cyclic property
of the trace, we obtain (see the derivation in Section \ref{section_proof_loss_csc_tr_G}):
\begin{align*}
h(\b{A})
&=
\operatorname{tr}\left[
\b{A}(\b{A}+\tau\b{I}_p)^{-1}
\right]\\
&=
\operatorname{tr}\left[
\b{\Lambda}
(\b{\Lambda}+\tau\b{I}_p)^{-1}
\right]\\
&=
\frac{\|\b{\mu}\|_2^2}
{\|\b{\mu}\|_2^2+\tau}.
\end{align*}
Since each centroid satisfies
$\|\b{c}_i\|_2\le1$, the triangle
inequality gives:
\begin{align*}
\|\b{\mu}\|_2
&=
\left\|
\frac{1}{b}\sum_{i=1}^{b}\b{c}_i
\right\|_2\le
\frac{1}{b}\sum_{i=1}^{b}\|\b{c}_i\|_2\le1.
\end{align*}
Thus:
\begin{equation*}
0\le\|\b{\mu}\|_2^2\le1.
\end{equation*}
The function $x/(x+\tau)$ is increasing
for $x\ge0$ because:
\begin{equation*}
\frac{d}{dx}\left(\frac{x}{x+\tau}\right)
=
\frac{\tau}{(x+\tau)^2}>0.
\end{equation*}
Consequently:
\begin{align*}
h(\b{A})
&=
\frac{\|\b{\mu}\|_2^2}
{\|\b{\mu}\|_2^2+\tau}\le
\frac{1}{1+\tau}=
\frac{r}{r+\kappa},
\end{align*}
where the last equality uses
$\tau=\kappa/r$.
Substituting this bound into Eq. (\ref{equation_hG_leq_hA_itau_tr_Sigma}) gives:
\begin{equation}\label{equation_hG_leq_r_r_plus_k_1_tau_tr_Sigma}
h(\b{G})
\le
\frac{r}{r+\kappa}
+
\frac{1}{\tau}\operatorname{tr}(\b{\Sigma}).
\end{equation}
According to Eqs. (\ref{equation_1_minus_l_csc_1_plus_k_r_hG}) and (\ref{equation_hG_leq_r_r_plus_k_1_tau_tr_Sigma}), we obtain:
\begin{align*}
1-\ell_{\mathrm{csc}}
&\le
\frac{1+\kappa}{r}
\left[
\frac{r}{r+\kappa}
+
\frac{1}{\tau}
\operatorname{tr}(\b{\Sigma})
\right]\\
&=
\frac{1+\kappa}{r+\kappa}
+
\frac{1+\kappa}{r\tau}
\operatorname{tr}(\b{\Sigma})\\
&=
\frac{1+\kappa}{r+\kappa}
+
\frac{1+\kappa}{\kappa}
\operatorname{tr}(\b{\Sigma}),
\end{align*}
since $r\tau=\kappa$.

We now rearrange this inequality
to obtain a lower bound on the
centered variance:
\begin{align*}
\frac{1+\kappa}{\kappa}
\operatorname{tr}(\b{\Sigma})
&\ge
1-\ell_{\mathrm{csc}}
-\frac{1+\kappa}{r+\kappa}\\
&=
\frac{r-1}{r+\kappa}
-\ell_{\mathrm{csc}}.
\end{align*}
Since $\kappa/(1+\kappa)>0$, we have:
\begin{equation*}
\operatorname{tr}(\b{\Sigma})
\ge
\frac{\kappa}{1+\kappa}
\left(
\frac{r-1}{r+\kappa}
-\ell_{\mathrm{csc}}
\right).
\end{equation*}
Furthermore, because $\b{\Sigma}$
is positive semidefinite, its trace
is always nonnegative:
\begin{equation*}
\operatorname{tr}(\b{\Sigma})\ge0.
\end{equation*}
Combining these two lower bounds,
we conclude:
\begin{equation}
\operatorname{tr}(\b{\Sigma})
\ge
\frac{\kappa}{1+\kappa}
\left[
\frac{r-1}{r+\kappa}
-\ell_{\mathrm{csc}}
\right]_+,
\end{equation}
where $[\cdot]_+=\max(\cdot,0)$ denotes
the positive-part operator.

Finally, suppose $r>1$ and:
\begin{equation*}
\ell_{\mathrm{csc}}
<
\frac{r-1}{r+\kappa}.
\end{equation*}
Under this condition, the expression
inside the positive-part operator is strictly positive:
\begin{equation*}
\frac{r-1}{r+\kappa}
-\ell_{\mathrm{csc}}>0.
\end{equation*}
Since $\kappa/(1+\kappa)>0$, the
centered-variance bound becomes:
\begin{equation}
\operatorname{tr}(\b{\Sigma})
\ge
\frac{\kappa}{1+\kappa}
\left(
\frac{r-1}{r+\kappa}
-\ell_{\mathrm{csc}}
\right)>0.
\end{equation}
Therefore, the centered covariance
cannot vanish, and the centroids
cannot all be identical.
This completes the proof.

\subsection{Proof of
Proposition~\ref{prop:csc_collapse_instability}}
\label{app:csc_collapse_instability_proof}

Consider the completely collapsed configuration:
\begin{equation*}
\b{u}_i=\b{v}_i=\b{w},
\qquad
\|\b{w}\|_2=1,
\qquad \forall i.
\end{equation*}
Since all centroids equal $\b{w}$:
\begin{equation*}
\b{G}=\b{w}\b{w}^\top.
\end{equation*}
Its eigenvalues are $1,0,\ldots,0$, so:
\begin{align}
\ell_{\mathrm{csc}}(0)
&=
1-\frac{1+\kappa}{r}
\frac{1}{1+\kappa/r}
\nonumber\\
&=
\frac{r-1}{r+\kappa}.
\label{eq:csc_collapse_original_loss}
\end{align}

We now construct an arbitrarily small perturbation
that increases spectral diversity.
Since $p\ge2$, choose a unit vector
$\b{e}\in\mathbb{R}^p$ satisfying:
\begin{equation*}
\b{e}^\top\b{w}=0,
\qquad
\|\b{e}\|_2=1.
\end{equation*}
For a sufficiently small $\theta>0$,
modify the representations of two instances:
\begin{align*}
\b{u}_1(\theta)
&=\b{v}_1(\theta)
=\cos\theta\,\b{w}
+\sin\theta\,\b{e},
\\
\b{u}_2(\theta)
&=\b{v}_2(\theta)
=\cos\theta\,\b{w}
-\sin\theta\,\b{e}.
\end{align*}
For all remaining instances, retain:
\begin{equation*}
\b{u}_i(\theta)=\b{v}_i(\theta)=\b{w},
\qquad i=3,\ldots,b.
\end{equation*}

All augmented views remain perfectly aligned.
Moreover, orthogonality of $\b{w}$ and $\b{e}$
ensures:
\begin{equation*}
\|\b{u}_i(\theta)\|_2
=
\|\b{v}_i(\theta)\|_2=1,
\qquad\forall i.
\end{equation*}
Because the views are identical, the centroids
equal their corresponding normalized representations.

The perturbed second-moment matrix is:
\begin{align}
\b{G}(\theta)
&=
\frac1b\sum_{i=1}^{b}
\b{c}_i(\theta)\b{c}_i(\theta)^\top
\nonumber\\
&=
\left(1-\frac{2\sin^2\theta}{b}\right)
\b{w}\b{w}^\top
+
\frac{2\sin^2\theta}{b}
\b{e}\b{e}^\top.
\label{eq:csc_perturbed_second_moment}
\end{align}

Define:
\begin{equation*}
x:=\frac{2\sin^2\theta}{b}.
\end{equation*}
For sufficiently small nonzero $\theta$,
we have $0<x<1$.
Since $\b{w}$ and $\b{e}$ are orthonormal,
the potentially nonzero eigenvalues become:
\begin{equation*}
\lambda_1=1-x,
\qquad
\lambda_2=x,
\end{equation*}
with all remaining eigenvalues equal to zero.
Notice that the total centroid energy is unchanged:
\begin{equation*}
t=\lambda_1+\lambda_2=1.
\end{equation*}
Thus, the perturbation increases spectral diversity
without reducing augmented-view agreement or
total centroid energy.

Recall:
\begin{equation*}
f_\tau(\lambda)=\frac{\lambda}{\lambda+\tau},
\qquad \tau=\frac{\kappa}{r}>0.
\end{equation*}
The perturbed CSC loss is:
\begin{equation*}
\ell_{\mathrm{csc}}(\theta)
=
1-\frac{1+\kappa}{r}
\left[
f_\tau(1-x)+f_\tau(x)
\right].
\end{equation*}
At the original collapsed configuration:
\begin{equation*}
\ell_{\mathrm{csc}}(0)
=
1-\frac{1+\kappa}{r}f_\tau(1).
\end{equation*}
Since $f_\tau$ is strictly concave
and $0<x<1$, we have:
\begin{equation*}
f_\tau(1-x)+f_\tau(x)
>
f_\tau(1)+f_\tau(0).
\end{equation*}
Using $f_\tau(0)=0$:
\begin{equation*}
f_\tau(1-x)+f_\tau(x)>f_\tau(1).
\end{equation*}
Therefore:
\begin{equation}
\ell_{\mathrm{csc}}(\theta)
<
\ell_{\mathrm{csc}}(0).
\label{eq:csc_collapse_descent}
\end{equation}
This proves that arbitrarily small perturbations
can strictly decrease the loss from the
completely collapsed configuration.

To establish the strict-saddle property,
we additionally observe that the gradient of
CSC with respect to $\b{C}$ is:
\begin{equation}
\nabla_{\b{C}}\ell_{\mathrm{csc}}
=
-\frac{2(1+\kappa)\tau}{rb}
(\b{G}+\tau\b{I}_p)^{-2}\b{C}.
\end{equation}
At the collapsed configuration, each column
of this gradient is proportional to $\b{w}$.
Consequently, its projection onto the tangent
space of each unit sphere vanishes. Hence, the
configuration is stationary under unit-norm
constraints.

However, along the constructed perturbation, we have:
\begin{equation*}
\left.
\frac{d^2\ell_{\mathrm{csc}}(\theta)}
{d\theta^2}
\right|_{\theta=0}
=
-\frac{4(1+\kappa)}{rb}
\left[
\frac1\tau
-\frac{\tau}{(1+\tau)^2}
\right]<0.
\end{equation*}
Thus, the stationary configuration admits a
direction of strictly negative curvature.
It is therefore a strict saddle, rather
than a local minimum, in the normalized
representation space.

\section{Minimal Loss Implementation}\label{section_minimal_loss_function}

The following PyTorch-style code specifies the mathematical computation. The recorded training implementation additionally checks finiteness and retries same-matrix CPU float64 SVD on solver nonconvergence. Neither an inverse nor a detached spectral approximation is required.

\begin{lstlisting}[
    language=Python,
    basicstyle=\ttfamily\footnotesize,
    breaklines=true,
    breakatwhitespace=false,
    columns=fullflexible,
    keepspaces=true,
    linewidth=\columnwidth,
    frame=single,
    framesep=5pt,
    rulecolor=\color{black}
]
import math
import torch
import torch.nn.functional as F

def csc_loss(z, z_prime, kappa=10.0):
    # z, z_prime: [batch, projection_dim]
    u = F.normalize(z, dim=1, eps=1e-6)
    v = F.normalize(z_prime, dim=1, eps=1e-6)
    c = (u + v) / 2  # no renormalization
    batch, dim = c.shape
    r = min(batch, dim)
    s = torch.linalg.svdvals(
        c / math.sqrt(batch)
    )
    eigenvalues = s.square()
    effective_dim = (
        eigenvalues /
        (eigenvalues + kappa / r)
    ).sum()
    return 1 - (1 + kappa) * effective_dim / r
\end{lstlisting}

\begin{table*}[t]
\centering
\caption{
\textbf{Additional multi-seed evaluation.}
Frozen ResNet-50 results averaged over
three random seeds. Classification
accuracies are reported in percentages.
All experiments use 30 training epochs.
}
\label{tab:supp_ablation}
\small
\begin{tabular}{llccc}
\toprule
Dataset & Method &
Linear Top-1 &
kNN Top-1 &
Alignment \\
\midrule

CIFAR-10
& CSC
& \textbf{78.07 $\pm$ 0.20}
& \textbf{75.91 $\pm$ 0.37}
& \textbf{0.8023 $\pm$ 0.0009} \\

& MMCR
& 77.34 $\pm$ 0.18
& 75.41 $\pm$ 0.08
& 0.7453 $\pm$ 0.0008 \\

\midrule

MNIST
& CSC
& \textbf{98.92 $\pm$ 0.17}
& \textbf{98.15 $\pm$ 0.09}
& \textbf{0.9933 $\pm$ 0.0003} \\

& MMCR
& 98.52 $\pm$ 0.06
& 97.21 $\pm$ 0.25
& 0.9904 $\pm$ 0.0004 \\

\midrule

CIFAR-100
& CSC
& 47.67 $\pm$ 0.42
& 41.33 $\pm$ 0.12
& \textbf{0.7833 $\pm$ 0.0012} \\

& CSC-Cov
& \textbf{47.92 $\pm$ 0.54}
& \textbf{41.36 $\pm$ 0.47}
& 0.7752 $\pm$ 0.0017 \\

\bottomrule
\end{tabular}
\end{table*}

\begin{table*}[t]
\centering
\caption{
\textbf{Sensitivity to the ridge parameter.}
CIFAR-100 results over three seeds
(mean $\pm$ standard deviation).
Alignment is measured using normalized
projector outputs; effective rank is
computed from centered frozen-encoder
covariance.
}
\label{tab:kappa_sensitivity}
\small
\begin{tabular}{ccccc}
\toprule
$\kappa$ &
Linear Top-1 (\%) &
kNN Top-1 (\%) &
Alignment &
Effective Rank \\
\midrule
1 &
47.08 $\pm$ 0.51 &
\textbf{41.44 $\pm$ 0.34} &
0.7103 $\pm$ 0.0013 &
\textbf{76.52 $\pm$ 0.83} \\

10 &
\textbf{47.67 $\pm$ 0.42} &
41.33 $\pm$ 0.12 &
0.7833 $\pm$ 0.0012 &
50.85 $\pm$ 0.69 \\

100 &
42.15 $\pm$ 0.72 &
33.97 $\pm$ 0.29 &
\textbf{0.8680 $\pm$ 0.0006} &
24.18 $\pm$ 0.22 \\
\bottomrule
\end{tabular}
\end{table*}

\section{Additional Experimental Results}
\label{app:additional_experiments}

We conduct additional controlled experiments
using ResNet-50 across three random seeds.
All experiments use 30 training epochs,
a physical batch size of 64, and gradient
accumulation over two steps.
These experiments use a shorter training
schedule than the main benchmarks and are
therefore reported separately.

\subsection{Multi-Seed Comparison with MMCR}

Table~\ref{tab:supp_ablation} presents additional
comparisons between CSC and MMCR on CIFAR-10
and MNIST.
On CIFAR-10, CSC achieves higher linear
Top-1 accuracy (78.07\% versus 77.34\%)
and kNN Top-1 accuracy (75.91\% versus
75.41\%). CSC also achieves stronger
positive-pair alignment.

Similar improvements are observed on MNIST,
where CSC achieves 98.92\% linear accuracy
and 98.15\% kNN accuracy, compared with
98.52\% and 97.21\% for MMCR.

These findings provide additional
multi-seed evidence supporting the
competitive representation quality
of the proposed objective.

\subsection{Effect of Covariance Centering}

We additionally compare CSC against a variant
that replaces the uncentered second-moment
matrix with the centered centroid covariance
matrix, while preserving the ridge-spectral
objective.

On CIFAR-100, CSC achieves slightly stronger
augmented-view alignment than CSC-Cov.
Downstream classification performance is
comparable between the two formulations,
indicating that covariance centering has
a limited effect under the evaluated
training configuration.

\subsection{Ridge-Parameter Sensitivity}
\label{app:kappa_sensitivity}

We evaluate the sensitivity of CSC to
$\kappa\in\{1,10,100\}$ on CIFAR-100
using ResNet-50. All configurations are
trained for 30 epochs with identical
settings and three random seeds.
Table~\ref{tab:kappa_sensitivity} reports
downstream classification accuracy,
positive-pair alignment, and effective rank.

\end{document}